\documentclass[lettersize,journal]{IEEEtran}
\usepackage{amsmath,amsfonts}
\usepackage{algorithmic}
\usepackage{algorithm}
\usepackage{array}
\usepackage[caption=false,font=normalsize,labelfont=sf,textfont=sf]{subfig}
\usepackage{textcomp}
\usepackage{stfloats}
\usepackage{url}
\usepackage{verbatim}
\usepackage{graphicx}
\usepackage{cite}

\usepackage{multirow}
\usepackage{booktabs}
\usepackage{makecell}

\usepackage{xcolor}
\usepackage{hyperref}
\usepackage[normalem]{ulem} 

\hypersetup{
    colorlinks=true,
    urlcolor=red,
    linkcolor=blue,
    citecolor=green
}

\newif\ifshowrevision
\showrevisionfalse

\newcommand{\add}[1]{\ifshowrevision\textcolor{blue}{#1}\else{#1}\fi}

\newcommand{\del}[1]{\ifshowrevision\textcolor{red}{\sout{#1}}\else{}\fi}

\newcommand{\replace}[2]{\ifshowrevision\textcolor{red}{\sout{#1}}~\textcolor{blue}{#2}\else{#2}\fi}

\newcommand{\rmark}[1]{\ifshowrevision{\footnotesize\textcolor{teal}{[#1]}}\fi}

\newenvironment{revised}{\ifshowrevision\bgroup\color{blue}\fi}{\ifshowrevision\egroup\fi}

\begin{document}





\title{AnimateAnyMesh++: A Flexible Feed-Forward Framework for High-Fidelity Text-Driven Mesh Animation}

\author{Zijie~Wu,
        Chaohui~Yu,
        Fan~Wang,
        and~Xiang~Bai\textsuperscript{*},~\IEEEmembership{Fellow,~IEEE}
\thanks{Zijie Wu is with the School of Artificial Intelligence and Automation, Huazhong University of Science and Technology, Wuhan, China, and also with DAMO Academy, Alibaba Group, Hangzhou, China (e-mail: zjw1031@hust.edu.cn).}
\thanks{Chaohui Yu is with DAMO Academy, Alibaba Group, Hangzhou, China, and also with Hupan Lab, Hangzhou, China (e-mail: huakun.ych@alibaba-inc.com).}
\thanks{Fan Wang is with DAMO Academy, Alibaba Group, Hangzhou, China (e-mail: fan.w@alibaba-inc.com).}
\thanks{Xiang Bai is with the School of Software Engineering, Huazhong University of Science and Technology, Wuhan, China (e-mail: xbai@hust.edu.cn).}
\thanks{\textsuperscript{*}Corresponding author: Xiang Bai.}
\thanks{A preliminary version of this work was presented at the IEEE/CVF International Conference on Computer Vision (ICCV), 2025 \cite{animateanymesh}}}

\markboth{IEEE Transactions on Pattern Analysis and Machine Intelligence}%
{Wu \MakeLowercase{\textit{et al.}}: AnimateAnyMesh++: A Flexible Feed-Forward Framework for High-Fidelity Text-Driven Mesh Animation}

\IEEEpubid{0000--0000/00\$00.00~\copyright~2026 IEEE}

\maketitle

\begin{abstract}
Recent advances in 4D content generation have attracted increasing attention, yet creating high-quality animated 3D models remains challenging due to the complexity of modeling spatio-temporal distributions and the scarcity of 4D training data. We present AnimateAnyMesh++, a feed-forward framework for text-driven animation of arbitrary 3D meshes with substantial upgrades in data, architecture, and generative capability. First, we expand the DyMesh-XL dataset by mining dynamic content from Objaverse-XL, increasing the number of unique identities from ~60K to ~300K and substantially broadening category and motion diversity. Second, we redesign DyMeshVAE-Flex with power-law topology-aware attention and vertex-normal–enhanced features, which significantly improves trajectory reconstruction, local geometry preservation, and mitigates trajectory-sticking artifacts. Third, we introduce architectural changes to both DyMeshVAE-Flex and the rectified-flow (RF) generator to support variable-length sequence training and generation, enabling longer animations while preserving reconstruction fidelity. Extensive experiments demonstrate that AnimateAnyMesh++ generates semantically accurate and temporally coherent mesh animations within seconds, surpassing prior approaches in quality and efficiency. The enlarged DyMesh-XL, the upgraded DyMeshVAE-Flex, and variable-length RF together deliver consistent gains across benchmarks and in-the-wild meshes. We will release code, models, and the expanded DyMesh-XL at \textbf{\textcolor{red}{\href{https://github.com/JarrentWu1031/AnimateAnyMesh-pp}{https://github.com/JarrentWu1031/AnimateAnyMesh-pp}}} upon acceptance of this manuscript to facilitate research in 4D content creation.
\end{abstract}

\begin{IEEEkeywords}
Mesh animation, 4D generation, Feed-forward, \del{Foundation model, }\rmark{R3-1}Rectified flow
\end{IEEEkeywords}  
\section{Introduction}
\label{Sec:intro}

\IEEEPARstart{T}{he} revolution in 3D content creation has fundamentally transformed domains like VR/AR and gaming. While generative modeling has established strong capabilities for synthesizing high-quality static 3D assets~\cite{instant3d,lgm,lrm,meshlrm}, extending these advances to 4D content generation presents formidable challenges due to the complexity of spatio-temporal distributions and the scarcity of 4D training data.


Existing 4D generation approaches generally fall into two categories: per-instance optimization methods~\cite{sc4d,consistent4d,4dfy,animate124,dreamgaussian4d} utilizing SDS~\cite{dreamfusion}, and multi-view dynamic video generation methods~\cite{animate3d,4diffusion}. The former suffers from high computational costs and inconsistency, while the latter relies on post-processing for 4D reconstruction, impeding real-time applications. Moreover, these methods, which typically adopt dynamic 3DGS~\cite{3dgs} or NeRF~\cite{nerf} as 4D representations, suffer from view discrepancies due to the lack of ground-truth 4D data, relying solely on multi-view rendering supervision.


\IEEEpubidadjcol

Given these limitations, we argue that dynamic meshes serve as an ideal representation for 4D creation. As the de facto standard in graphics pipelines, meshes offer superior rendering efficiency and naturally decouple geometry from motion. Furthermore, traditional rigging pipelines, while dominant, assume a low-dimensional articulation space and require labor-intensive weight painting, limiting their applicability to generic or topologically complex assets. In contrast, we pursue a rig-free, \textbf{vertex-based} mesh animation pipeline that directly models per-vertex trajectories. This avoids discretization gaps between skeleton kinematics and surface deformation, and natively handles arbitrary topologies and non-rigid effects.

To this end, we introduce \textbf{AnimateAnyMesh++}, a feed-forward framework for text-driven universal mesh animation. At its core, we introduce \textbf{DyMeshVAE-Flex}, a novel VAE~\cite{vae} architecture tailored for dynamic mesh sequences, aiming to compress and reconstruct the trajectory of each vertex. To address the inherent challenges of modeling complex spatio-temporal mesh deformations, DyMeshVAE-Flex is built upon three meticulously designed components, each driven by specific motivations to overcome limitations in existing sequence modeling:
First, to break the constraint of fixed-length sequence modeling, we introduce a chunk-based processing strategy coupled with \textit{Time-Dependent Gradient Weighting (TDGW)}. By segmenting relative trajectories into overlapping chunks and reassembling them using TDGW, this design maintains continuous temporal information flow and ensures seamless transitions between chunks. Consequently, it natively supports variable-length sequence training and generation.
Second, to preserve fine-grained local geometry and prevent the common artifact of trajectory entanglement, we propose \textit{Power-Law Topology-Aware (PLTA) Attention} alongside \textit{Vertex Normal Injection}. Specifically, PLTA Attention is designed to exponentially expand the local receptive field of each vertex. Combined with explicitly injected normal features, the vertices are enriched with comprehensive neighborhood information, ensuring that fine-grained local geometry is strictly preserved during complex motions.
Third, to effectively mitigate confusion between geometry and motion modeling, we introduce \textit{Sync Attention}. Conventional full attention mechanisms often suffer from information aliasing between spatial and temporal dimensions. Sync Attention addresses this by exploiting the inherent local cooperativity prior of mesh motion—using the attention maps derived from vertex features to guide the projection of corresponding trajectory features. This methodology ensures complete isolation between vertex and trajectory features, thereby preventing mutual interference.

To bridge the gap between text descriptions and mesh animations, we propose \textbf{Shape-Guided Text-to-Trajectory (SGTT) Model}. This model is composed of alternating spatial-temporal self-attention and \replace{text cross-attention}{MMDiT-style~\cite{sd3} joint self-attention with text tokens}\rmark{R1-1}, and leverages a Rectified Flow-based~\cite{rf} training strategy in the compressed latent space. By learning the conditional distribution of relative trajectories given text prompts and initial mesh features, our approach enables the generation of smooth and realistic animations.
Furthermore, as a guarantee of AnimateAnyMesh++'s performance, we introduce a large-scale dynamic mesh dataset called \textbf{DyMesh-XL}. This dataset integrates diverse data sources such as Objaverse-XL~\cite{objaverse-xl}, Objaverse-1.0~\cite{objaverse}, AMASS~\cite{AMASS}, and DeformingThings4D~\cite{dt4d}. Through meticulous processes like vertex trajectory extraction, motion filtering, and high-quality caption generation, DyMesh-XL ultimately comprises over \textbf{4M} segments of dynamic mesh data, spanning 16/32/64 frames. This provides a solid foundation for training and evaluation in the field of mesh animation.

This paper is an extended version of our conference paper~\cite{animateanymesh}, published in ICCV 2025. \begin{revised}While we deliberately retain the overall feed-forward paradigm of the conference version -- compressing vertex trajectories into a compact latent space and generating them via a text-conditioned rectified-flow model, since this paradigm was already validated to be efficient and effective -- each of its core components is substantially redesigned to remove specific limitations that prevented the conference version from scaling in data, sequence length, and geometric fidelity. Concretely, we make the following new contributions:\end{revised}\rmark{R3-2}

\begin{enumerate}
    \item We introduce DyMeshVAE-Flex, which achieves high-fidelity, variable-length mesh sequence compression through chunk-based compression and Time-Dependent Gradient Weighting (TDGW).
    \item We propose Power-Law Topology-Aware (PLTA) Attention and Vertex Normal Injection, which significantly enhance the local receptive field of vertex features, leading to improved local geometry preservation and the avoidance of trajectory sticking issues.
    \item We introduce the DyMesh-XL dataset, which significantly augments data from Objaverse-XL and includes extensive long-sequence data.
    \item The proposed AnimateAnyMesh++ surpasses all existing text-driven mesh animation methods, including our conference version, in both performance and efficiency, significantly broadening the practical scope of feed-forward 4D generation.
\end{enumerate}  
\section{Related Works}

\noindent\textbf{3D Generation.}
Early approaches~\cite{dreamfusion,dream3d,sjc,magic3d,prolificdreamer,dreamcraft3d,pointsto3d,make-it-3d,dreamgaussian,gsgen,gaussiandreamer} for 3D generation leverage CLIP~\cite{clip} score or Score Distillation Sampling (SDS)~\cite{dreamfusion} to distill geometric priors from pre-trained 2D generative models~\cite{sd, imagen}. However, due to the inherent lack of 3D information, these methods often suffer from view discrepancy issues and require time-consuming per-scene optimization, significantly limiting their practical applications.
To address these limitations, some methods~\cite{zero123,imagedream,syncdreamer,zero123++,richdreamer} try to fine-tune 2D generative models using multi-view renderings of 3D assets~\cite{objaverse,objaverse-xl}, thereby enhancing view consistency in the generated 3D content. Nevertheless, these approaches still require per-instance reconstruction after obtaining multi-view outputs.
In contrast, feed-forward methods~\cite{instant3d,lgm,12345,12345++,lrm,meshlrm,instantmesh,grm,gs-lrm,zhang2025lpm} directly generate 3D representations~\cite{3dgs,nerf} and optimize the networks through multi-view rendering supervision, enabling rapid 3D asset generation within seconds for given prompts. This end-to-end approach circumvents the need for expensive post-processing optimization.
More recent feed-forward 3D generation works~\cite{hunyuan1.0, hunyuan2.0, trellis, triposg, sparseflex, hi3dgen, direct3d} have further elevated the fidelity of the generated 3D assets to an application level, owing to their sophisticated tokenizers.

Inspired by the evolution of 3D generation techniques, we pioneer a feed-forward 4D architecture for universal mesh animation. Our approach enables rapid mesh animation in a few seconds without per-scene optimization or reconstruction.

\noindent\textbf{4D Generation.}
Following the evolution path of 3D generation, early 4D generation approaches~\cite{sc4d,consistent4d,4dfy,animate124,dreamgaussian4d,alignyg,mav3d,4dgen,efficient4d,stag4d} attempt to distill 4D priors from 2D/3D/video generative models. Compared to 3D generation, these distillation approaches for 4D content creation not only demand significantly more computing power and longer optimization time to generate a single scene, but also tend to produce more noticeable spatio-temporal artifacts.
Some works~\cite{animate3d,4diffusion,diffusion4d,l4gm,vividzoo} have explored finetuning 3D/video generative models~\cite{mvdream,animatediff,zeroscope,modelscope} with 4D data to synthesize multi-view dynamic videos, aiming to accelerate 4D generation and improve spatio-temporal consistency. \replace{Nevertheless, these approaches still rely heavily on per-scene 4D reconstruction, and the generated objects exhibit spatial discrepancies due to the lack of true 4D training data.}{For example, Animate3D~\cite{animate3d} and L4GM~\cite{l4gm} respectively rely on per-instance optimization and monocular-video-conditioned feed-forward prediction to obtain 4D representations from such synthesized videos, and both still suffer from limited motion amplitude and geometric inconsistency due to the lack of direct 4D supervision (Sec.~\ref{sec:exp}).}\rmark{R2-1} While some methods~\cite{motion2vecsets,humanmdm,motiondiffuse,realistichumanmt,physdiff} have achieved efficient 4D generation in specific categories (e.g., human bodies) through parametric models~\cite{smpl} and modality-specific data~\cite{dt4d,AMASS}, \add{these methods are constrained to a fixed skeletal parameterization and cannot generalize to arbitrary mesh topology.}\rmark{R2-1} To the best of our knowledge, direct feed-forward mesh animation for general categories remains unexplored. \add{We note that an analogous animation task has also been studied in the 2D/video domain~\cite{motioneditor,motionfollower,stableanimator,stableanimatorpp}, where pixel-space videos or images are animated given driving poses; our task can be viewed as its 3D counterpart, directly generating text-driven vertex trajectories for meshes of arbitrary topology.}\rmark{R3-ref}

\del{In this work, we propose a novel framework, AnimateAnyMesh++, which facilitates efficient, text-controlled, feed-forward mesh animation. This framework is composed of two key components: DyMeshVAE-Flex, responsible for the high-quality compression and reconstruction of dynamic mesh vertex trajectories, and a Shape-Guided Text-to-Trajectory (SGTT) model that learns the distribution of these vertex trajectories. By integrating the training and sampling strategies of Rectified Flow, our method achieves feed-forward, text-driven animation for meshes of arbitrary topology and category at test time.}\add{Motivated by these limitations, we propose AnimateAnyMesh++, detailed in Sec.~\ref{sec:dymeshvae-flex}-\ref{sec:sgtt}.}\rmark{R1-5} \replace{This work marks a solid step towards universal 4D generation.}{This provides a practical and extensible framework for high-fidelity, text-driven animation of arbitrary-topology meshes.}\rmark{R3-1}

\section{DyMeshVAE-Flex}
\label{sec:dymeshvae-flex}
\noindent \del{Existing mesh animation methods often rely on per-instance optimization or are category-specific. To address this, we propose AnimateAnyMesh++, a feed-forward framework for text-driven universal mesh animation. It}\add{AnimateAnyMesh++} consists of two components: DyMeshVAE-Flex and a Shape-Guided Text-to-Trajectory (SGTT) model. \del{In this chapter, we meticulously detail the architectural components of DyMeshVAE-Flex, presenting its encoder, and decoder in Sec.~\ref{sec:dvae_enc}, and Sec.~\ref{sec:dvae_dec}, respectively.}\add{This section details the architecture of DyMeshVAE-Flex, covering its encoder (Sec.~\ref{sec:dvae_enc}) and decoder (Sec.~\ref{sec:dvae_dec}).}\rmark{R1-5}

\begin{revised}
\subsection{Problem Formulation}
\label{sec:formulation}
\noindent Given a static 3D mesh $(F, V_0)$, with face connectivity $F\in\mathbb{R}^{M\times3}$ and initial vertex positions $V_0\in\mathbb{R}^{N\times3}$, and a text prompt $c$ describing the desired motion, our goal is to synthesize a temporally coherent, text-aligned animation of $T$ frames. We represent this animation as per-vertex relative trajectories $V_T\in\mathbb{R}^{N\times(T\cdot3)}$ satisfying $V^t=V_0^t+V_T^t$, so that $\{F, V_0, V_T\}$ fully specifies the animated dynamic mesh sequence $D$.

We factorize this task into the two components introduced above: DyMeshVAE-Flex learns a compact, variable-length latent space for the trajectories $V_T$ via an encoder-decoder pair, independent of any text condition (detailed in the remainder of this section); SGTT (Sec.~\ref{sec:sgtt}) then learns the distribution of this latent representation conditioned on $(V_0, c)$ via Rectified Flow~\cite{rf}. At inference time, we sample a latent from SGTT given $(V_0, c)$ and decode it with DyMeshVAE-Flex to obtain the final animation.
\end{revised}\rmark{R2-2}

\begin{figure*}[t]
  \centering
   \includegraphics[width=1.0\linewidth]{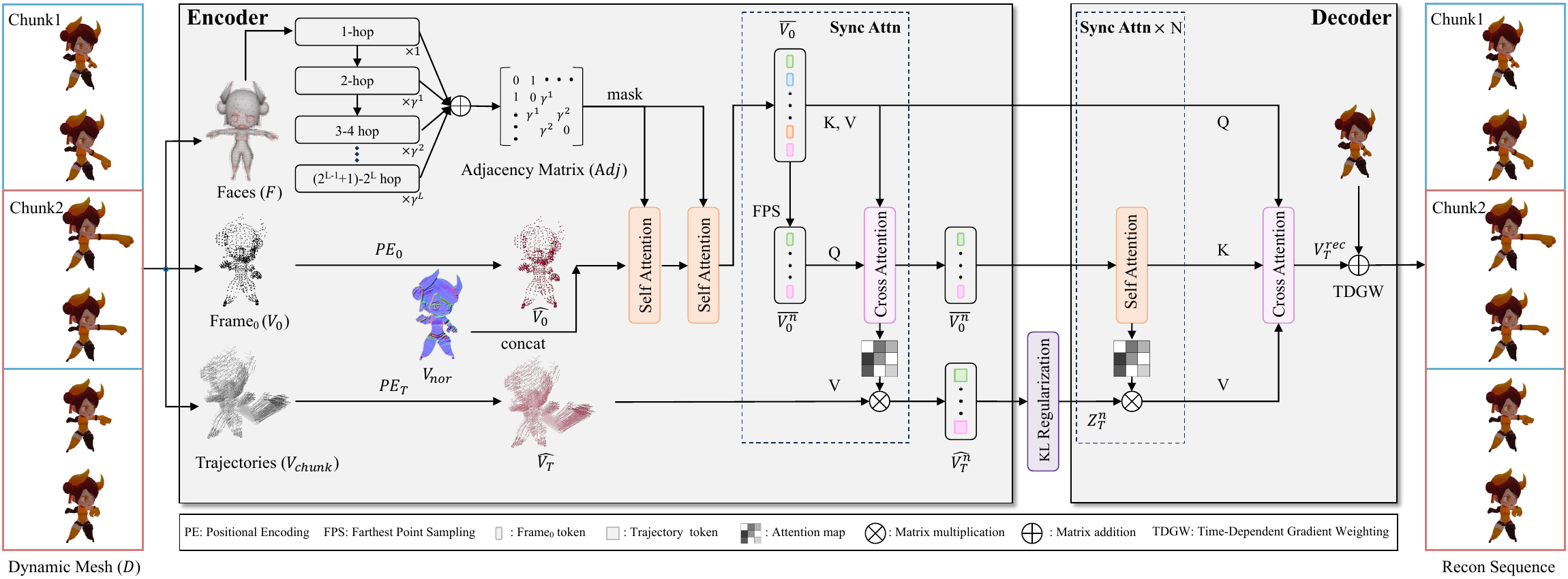}
   \caption{Illustration of our proposed \textbf{DyMeshVAE-Flex}. A long dynamic mesh sequence \(D\) is first segmented into overlapping chunks. The initial vertices \(V_0\), static faces \(F\), and relative trajectories \(V_T\) of each chunk are mapped into a decoupled latent space \(\{\overline{V_0^n}, \widehat{V_T^n}\}\) via the \textbf{Encoder}, utilizing trajectory decomposition and topology-aware attention. The \textbf{Decoder} then reconstructs the relative trajectories \(V^{rec}_T\) from the latents. Finally, the reconstructed chunks are seamlessly fused into a smooth, full-length dynamic mesh using our Time-Dependent Gradient Weighting (TDGW).}
   \label{fig:dvae}
\end{figure*}

\subsection{DyMeshVAE-Flex Encoder}
\label{sec:dvae_enc}
\noindent As shown in Fig.~\ref{fig:dvae}, given a dynamic mesh sequence $D\subset \{F\in\mathbb{R}^{M\times3}, V\in\mathbb{R}^{T\times N\times 3}\}$, we first disentangle the vertices sequence $V$ into the initial frame vertices $V_0\in\mathbb{R}^{N\times 3}$ and the relative trajectories $V_T\in\mathbb{R}^{N\times (T\cdot 3)}$, which satisfy: $V^t = V_0^t + V_T^t$,
where $t$ stands for the index of the time sequence. 

We decompose the vertex sequence $V$ into initial positions $V_0$ and relative trajectories $V_T$ based on our empirical observation that such decomposition leads to the disentanglement of shape and motion, while yielding a motion distribution that better approximates a zero-mean normal distribution.
Following this decomposition, each trajectory is represented as a combination of its initial vertex position and the subsequent temporal offsets. The initial position serves as a spatial identifier, while the temporal offsets become our modeling target. Then, we split $V_T$ into overlapping chunks $V_{chunk}$ to support the compression of variable-length sequences:

\begin{equation}
\begin{aligned}
  &V_{chunk_0} = concat(\mathbf{0},V_T[:L_S]), \\
  &V_{chunk_i} = V_T[(i+1)\cdot L_S-L_C:(i+1)\cdot L_S],
  \label{eq:v_sep}
\end{aligned}
\end{equation}
where $\mathbf{0}$ denotes the padded zeros. $L_C$ is the chunk size including the overlap region, $L_S$ is the length of the non-overlapping region within each chunk. Through this overlapping segmentation, each chunk inherently incorporates a portion of the preceding chunk's trailing information, thereby establishing the foundation for a smooth transition between chunks during the reconstruction phase.

\begin{figure}[t]
  \centering
   \includegraphics[width=0.85\linewidth]{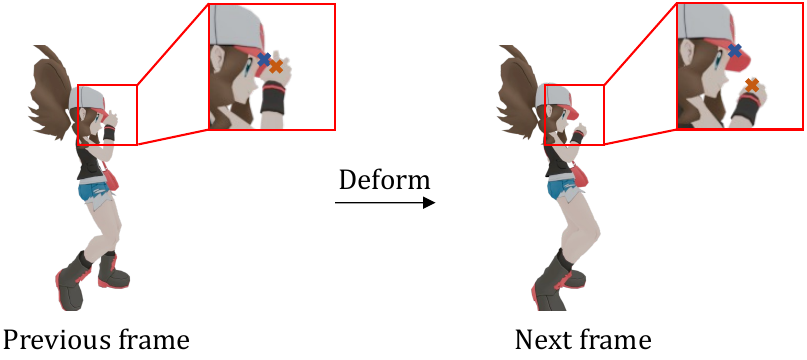}
   \caption{Demonstration of divergent trajectories for nearby mesh vertices.}
   \label{fig:traj_stuck}
\end{figure}

To prevent adhesion effects and enhance trajectory reconstruction stability (as demonstrated in Fig.~\ref{fig:traj_stuck}), we employ positional encoding schemes for $V_0$ and $V_T$, resulting in encoded features $\widehat{V_0}$ and $\widehat{V_T}$.

\noindent\textbf{Vertex Normal Injection}.
In addition to vertex coordinates, vertex normals provide rich local geometric cues. This is highly beneficial for the network when learning the fine-grained local deformation (or motion) of an object. Specifically, for a vertex $v_0$ in the initial frame, let $f_1, f_2, \dots, f_m$ be the set of faces incident to it. The normal vector for this vertex is calculated as:

\begin{equation}
  v_{nor}^0 = \frac{\sum_{i=1}^{m} s_i n_i}{\left\| \sum_{i=1}^{m} s_i n_i \right\|},
  \label{eq:v_norm}
\end{equation}
where $s_i, n_i$ are the surface area and normal of $f_i$. Once the normal for each vertex is computed, we separately project the vertex normal features and the coordinate features through linear layers and then concatenate them, providing a more discriminative representation for each vertex:

\begin{equation}
  \widehat{V_0} = Concat(\widehat{V_0},~V_{nor}).
  \label{eq:v_norm}
\end{equation}

\noindent\textbf{Power-Law Topology-Aware (PLTA) Attention}.
Due to intrinsic local rigidity, vertices in proximate regions exhibit strong motion correlations. However, relying solely on spatial coordinates causes semantic ambiguity (e.g., spatially close but semantically distinct parts like a hand and a hat, as in Fig.~\ref{fig:traj_stuck}), leading to trajectory entanglement. To resolve this, we must aggregate topological context. 

Specifically, given the initial mesh's face matrix $F$, we first derive the 1-hop adjacency matrix $H_1$, which encodes the direct connectivity between vertices. By utilizing $H_1$ as an attention mask, we employ self-attention to aggregate the features of directly connected neighbors into the target vertex, thereby enhancing its feature discriminability.

However, for complex topologies (i.e., high-resolution meshes), aggregating only the 1-hop neighborhood is insufficient. Since direct neighbors may be spatially very close, the aggregated information still lacks the necessary context to differentiate the target vertex from others belonging to spatially proximate but semantically distinct regions.

While stacking multiple 1-hop self-attention layers could theoretically expand the receptive field, it incurs massive computational overhead and causes feature over-smoothing. Conversely, recursively computing exact multi-hop matrices is prohibitively slow and memory-intensive for high-resolution meshes. To efficiently capture extensive topological context, we propose computing hop-band matrices with an exponentially growing receptive field. Instead of calculating exact \(n\)-hop connections, we iteratively compute connectivity bands:
\begin{revised}\rmark{R3-m1,R3-m2}
\begin{equation}
  \begin{aligned}
    C_{2^{L_{hop}}} &= \text{Boolean}(C_{2^{L_{hop}-1}} \cdot C_{2^{L_{hop}-1}}) \lor C_{2^{L_{hop}-1}}, \\
    Band_{L_{hop}} &= C_{2^{L_{hop}}} \land \neg C_{2^{L_{hop}-1}}, \\
    Adj &= \sum_{i=0}^{L_{hop}} \gamma_{hop}^i \cdot Band_i,
  \end{aligned}
  \label{eq:hop_band}
\end{equation}
where \(C_1 = H_1\) is the initial connectivity matrix. The hop band matrix \(Band_{L_{hop}}\) represents the newly discovered neighborhood band at scale \(L_{hop}\). We set \(L_{hop}=4\) and the decay coefficient \(\gamma_{hop}=0.5\). This approach yields an exponentially expanded receptive field using only simple matrix operations, perfectly aligning with the intuition of local perception: closer neighborhoods require finer feature differentiation, while distant ones only need coarse contextual information.
\end{revised}

Finally, we use the resulting $Adj$ matrix as an attention mask to compute self-attention for the features of all mesh vertices:

\begin{equation}
  \overline{V_0} = \text{Softmax}\left( \frac{\widehat{V_0}\cdot \widehat{V_0}^T}{\sqrt{d_k}} + \text{log}(Adj+\epsilon)\right) \widehat{V_0} + \widehat{V_0} 
\end{equation}
where $d_k$ denotes the channel dimension of the projected space, and $\epsilon$ is a small hyperparameter used to ensure that every element in the $\log$ calculation is positive.

We refer to the aforementioned attention mechanism for aggregating neighborhood information as Power-Law Topology-Aware (PLTA) attention. \replace{We stack two layers of PLTA attention to further expand the local receptive field. In our default setting ($L=4$), our theoretical local receptive field per vertex can reach $(2^{4-1})^2 = 64$.}{As derived in Eq.~\ref{eq:hop_band}, a single PLTA attention layer aggregates information from all vertices within $2^{L_{hop}}$ hops, i.e., $16$ hops under our default setting ($L_{hop}=4$). We further stack two layers of PLTA attention to additively extend this effective receptive field to $2\times 2^{L_{hop}}=32$ hops.}\rmark{R3-6}

\noindent\textbf{Sync Attention}.
Following the acquisition of the topology-aware vertex features $\overline{V_0}$, our subsequent objective is to explicitly model the inherent correlation observed in the trajectories of locally adjacent vertices. To achieve this, we introduce the \replace{Synchro (Sync)}{Sync}\rmark{R3-m3} Attention mechanism, designed to ensure that the aggregation and subsequent separation of trajectories are governed by the local topology-aware vertex features. Specifically, we initially employ Farthest Point Sampling (FPS) to select $n$ representative features, denoted $\overline{V_0^n}$, from $\overline{V_0}$. \add{Note that FPS is consistently applied to the topology-aware vertex features $\overline{V_0}$ (i.e., the output of PLTA attention, Eq.~\ref{eq:hop_band}) rather than the raw vertex coordinates $V_0$; since $\overline{V_0}$ remains well-defined regardless of the specific PLTA hyperparameters, this holds for every encoder configuration, including all ablation variants in Sec.~\ref{sec:exp}.}\rmark{R3-m4} Subsequently, we perform a cross-attention computation, utilizing $\overline{V_0^n}$ as the query, and $\overline{V_0}$ as both the key and value. The resulting attention map from this process is then directly applied to the corresponding $\hat{V_T}$ to yield the aggregated trajectories feature $\hat{V_T^n}$, which are associated with the sampled features $\overline{V_0^n}$:

\begin{equation}
\begin{aligned}
  \overline{V_0^n} = \text{Softmax}\left( \frac{\overline{V_0^n}\cdot \overline{V_0}^T}{\sqrt{d_k}  }\right) \overline{V_0} + \overline{V_0^n}, \\
  \widehat{V_T^n} = \text{Softmax}\left( \frac{\overline{V_0^n}\cdot \overline{V_0}^T}{\sqrt{d_k}  }\right) \widehat{V_T} + \widehat{V_T^n}.
  \label{eq:enc_ca_vt}
\end{aligned}
\end{equation}

Consequently, both $\overline{V_0^n}$ and $\widehat{V_T^n}$ are concurrently enriched with comprehensive local and global information. Moreover, the aggregation of trajectories, being predicated on the local correlation among vertices, inherently capitalizes on the principle of local object rigidity. This mechanism substantially mitigates the complexity of the learning complex motion distributions for the network.




After encoding, the dynamic mesh sequence is compressed into latents \(\{\overline{V_0^n}, \widehat{V_T^n}\}\). Since our focus is animating existing meshes, we only model the distribution of the relative trajectories \(\widehat{V_T^n}\). Following standard latent diffusion frameworks~\cite{sd,3dshape2vecset}, we apply KL-regularization to modulate feature diversity. Specifically, we linearly project \(\widehat{V_T^n}\) to predict the mean \(\mu_T^n\) and variance \(\sigma_T^n\), sample the latent \(Z_T^n\) via the standard reparameterization trick \(Z_T^n = \mu_T^n + \sigma_T^n \cdot \epsilon\) (where \(\epsilon \sim \mathcal{N}(0,1)\)), and compute the standard KL divergence loss \(L_{kl}\) to regularize the latent space towards a standard normal distribution.

\subsection{DyMeshVAE-Flex Decoder}
\label{sec:dvae_dec}
\noindent To leverage motion correlations among vertices with similar local structures, we process the topology-aware features \(\overline{V_0^n}\) and latents \(Z_T^n\) through \(K\) cascaded Sync Attention blocks. Specifically, we compute self-attention maps from \(\overline{V_0^n}\) and apply them to update both \(\overline{V_0^n}\) and \(Z_T^n\). Subsequently, we decode the trajectory chunks \(V_{chunk}^{rec}\) via cross-attention, utilizing the initial mesh features \(\overline{V_0}\) as queries, and \(\overline{V_0^n}\) and \(Z_T^n\) as keys and values, respectively.

\noindent\textbf{Time-Dependent Gradient Weighting (TDGW).}
Concatenating independently decoded trajectory chunks often introduces temporal discontinuities (i.e., "jumps") at boundaries. Since our encoding process incorporates overlapping intervals between adjacent chunks, we ensure seamless transitions by applying Time-Dependent Gradient Weighting (TDGW) within these overlapping regions\add{ of length \(L_O = L_C - L_S\)}\rmark{R3-m1}. For \replace{an}{such an} overlap, the blended trajectory at step \(t \in \{0, \dots, L_O-1\}\) is computed via linear interpolation:
\begin{revised}\rmark{R3-7}
\begin{equation}
  V_{T}^{rec}[t_{ov}] = (1 - W(t)) \cdot V_{chunk_i}^{rec}[t_{ov}] + W(t) \cdot V_{chunk_{i+1}}^{rec}[t],
\end{equation}
where \(t_{ov} = L_C - L_O + t\) is the corresponding time index in the preceding chunk \(i\) (with chunk size \(L_C\)), and the weight \(W(t) = (t+1)/(L_O+1)\) gradually increases from near \(0\) to near \(1\). Consequently, the contribution of the preceding chunk \(i\), weighted by \(1-W(t)\), gradually decays as the succeeding chunk \(i+1\) progressively takes over, yielding a smooth, continuous full-length relative trajectory \(V_T^{rec}\).
\end{revised}

Finally, the absolute vertex positions of the reconstructed dynamic mesh are recovered by transposing \(V_T^{rec}\) and adding the initial positions \(V_0\). The entire VAE is trained end-to-end using a weighted sum of the standard MSE reconstruction loss \(L_{rec}\) and the KL divergence loss: \(L_{dvae} = L_{rec} + \eta \cdot L_{kl}\), where we set \(\eta = 10^{-6}\) by default.

\section{Shape-Guided Text-to-Trajectory Model}
\label{sec:sgtt}

\begin{figure}[t]
  \centering
   \includegraphics[width=0.75\linewidth]{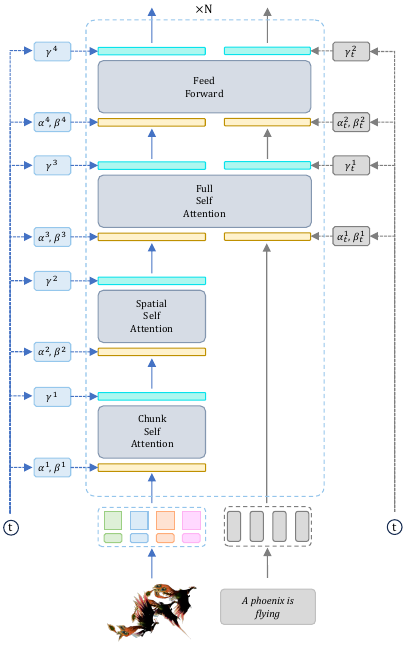}
   \caption{The architecture of the Shape-Guided Text-to-Trajectory (SGTT) Model. The dynamic mesh and text features are extracted and fed into a stack of \replace{$N$}{$N_{sgtt}$}\rmark{R3-m1} transformer blocks. Both streams are modulated by the diffusion time step $t$ via scale, shift, and gate parameters \replace{($\alpha, \beta, \gamma$)}{($\alpha, \beta, \delta$)}\rmark{R3-m2} at various stages.}
   \label{fig:sgtt}
\end{figure}

\begin{figure*}[t]
  \centering
   \includegraphics[width=1.0\linewidth]{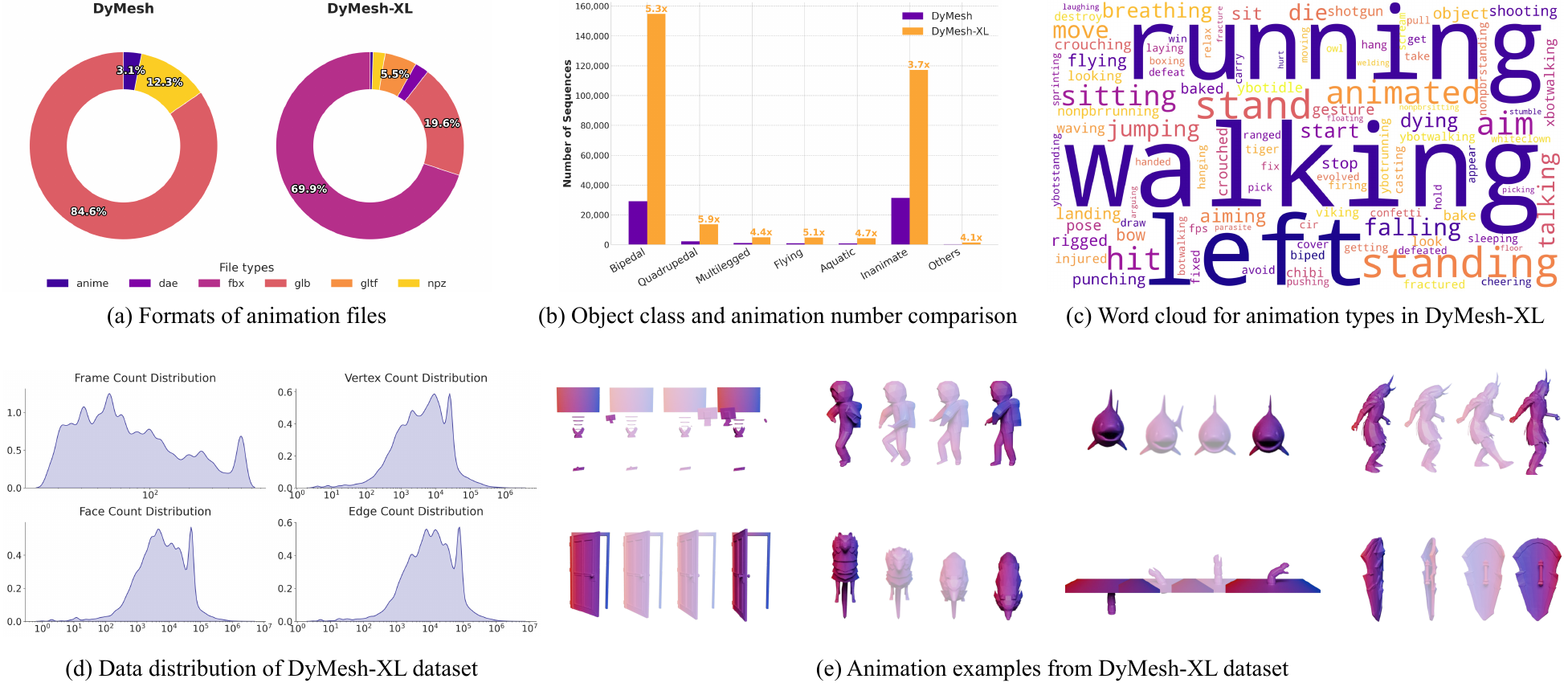}
   \caption{Comprehensive overview of the DyMesh-XL dataset. (a) Comparison of animation file formats in DyMesh and DyMesh-XL. (b) Quantitative comparison of object classes, highlighting the significant increase in animation sequences in DyMesh-XL. (c) A word cloud of the top 100 most frequent action verbs in DyMesh-XL, illustrating the dataset's action diversity. (d) Statistical distributions of frame, vertex, face, and edge counts for the data in DyMesh-XL. (e) A gallery of rendered animation examples from DyMesh-XL, showcasing the variety of objects and motions.}
   \label{fig:dymesh-xl}
\end{figure*}

\noindent For text-driven mesh animation, our model learns to estimate the posterior distribution of relative trajectories conditioned on both the initial mesh and textual prompts.
As shown in Fig.~\ref{fig:sgtt}, we constructed a trajectory generation model based on the DiT~\cite{dit} architecture, which is termed as the Shape-Guided Text-to-Trajectory (SGTT) Model. 

\subsection{Architecture}
\noindent The proposed SGTT network consists of \(\replace{N}{N_{sgtt}}\rmark{R3-m1}\) identical Transformer blocks\add{, using a distinct symbol from the vertex count $N$ in Sec.~\ref{sec:dymeshvae-flex} to avoid ambiguity}\rmark{R3-m1}. Specifically, as described in Sec.~\ref{sec:dymeshvae-flex}, the concatenated latent \(Z^n \in \mathbb{R}^{num_c \times n \times c}\) (where \(num_c\) denotes the number of chunks, \add{$n$ is the fixed number of latent tokens per chunk obtained via cross-attention pooling in the DyMeshVAE-Flex encoder (independent of the vertex count $N$), and $c$ is the latent channel dimension}\rmark{R3-m1}) serves as the input. Before each attention layer, we apply AdaLN-Zero~\cite{cogvideox} to modulate the features using scale and shift ($\alpha, \beta$) parameters derived from the diffusion timestep \(t\). 

The latent is sequentially processed by four main components: (1) a chunk-wise self-attention layer equipped with Rotary Position Embedding (RoPE)~\cite{rope} to capture temporal dynamics; (2) a spatial-wise self-attention layer to model intra-chunk structural relations; (3) \replace{a full cross-attention layer where the flattened 4D features jointly query the text embeddings}{a full self-attention layer, following the MMDiT paradigm~\cite{sd3}, where the flattened 4D features are concatenated with the text embeddings and jointly attend to one another}\rmark{R1-1}; and (4) a feed-forward network (FFN). The output of each attention layer is scaled by a timestep-dependent gating parameter \add{$\delta$}\rmark{R3-m2} and added residually. 

This alternating spatial-temporal attention scheme significantly reduces computational overhead and mitigates interference among misaligned chunks. Furthermore, the decoupled reparameterization and gating design enables adaptive modulation across different denoising stages, establishing a robust foundation for modeling text-conditioned vertex trajectories.

\subsection{Diffusion Pipeline}
\label{diff_pipe}
\noindent\textbf{Training.}
Following the Rectified Flow (RF)~\cite{rf} paradigm, we aim to minimize the mean square error between the predicted and ground truth velocity field. Since \(\overline{V_0^n}\) is fixed, our model focuses solely on learning the distribution of the relative trajectory features \(Z_T^n\). The noisy latent is constructed as \(\tilde{Z}_T^n = (1-t)Z_T^n + t\epsilon\), where \(\epsilon \sim \mathcal{N}(0,1)\) and the timestep \(t\) is sampled via \(t = 1 - (1 / (\tan(\frac{\pi}{2}u) + 1))\) with \(u \sim \mathcal{U}(0,1)\). The optimization objective is formulated as the standard flow matching loss:
\begin{equation}
  L_{rf} = \mathbb{E}_{\overline{V_0^n}, C_{text}, \epsilon, t}\left \|  v_\theta(\tilde{Z}_T^n;\overline{V_0^n}, C_{text}) - (Z_T^n - \epsilon) \right \| _2^2,
  \label{eq:rf_loss}
\end{equation}
where \(v_\theta\) represents the SGTT backbone.

\noindent\textbf{Inference.}
During inference, we employ standard flow-based ODE sampling to generate \(Z_T^n\) from pure noise. To enhance text-alignment, we incorporate Classifier-Free Guidance (CFG) to modify the predicted velocity field, setting the guidance scale to \(\zeta=3.0\) by default. Finally, we feed both the sampled features \(Z_T^n\) and the topology-aware vertex features \(\overline{V_0^n}\) into the decoder to reconstruct the relative trajectories, which are then added to the initial mesh to generate the final animated sequence. 

\section{DyMesh-XL Dataset}
\label{sec:dymeshxl}

\noindent To train our AnimateAnyMesh++ framework, we introduce \textbf{DyMesh-XL}, a large-scale 4D mesh dataset aggregating diverse sequences from Objaverse-XL~\cite{objaverse-xl}, Objaverse~\cite{sketchfab,objaverse}, AMASS~\cite{AMASS}, and DT4D~\cite{dt4d}. 

\noindent\textbf{Data Processing.} We extract animation files across various formats (e.g., \texttt{.fbx, .glb}) and convert them into unified vertex trajectories and triangular meshes, capping lengths between 16 and 200 frames. Each sequence is parameterized as \(D = \{F \in \mathbb{R}^{M \times 3}, V \in \mathbb{R}^{T \times N \times 3}\}\), where \(M\), \(T\), and \(N\) denote the face count, temporal length, and vertex count, respectively. To optimize data and support the topology-aware encoding of DyMeshVAE-Flex, we merge duplicate vertices and update face connectivity accordingly. 

\noindent\textbf{Augmentation and Filtering.} The sequences undergo overlapping temporal slicing with window sizes \(T \in \{16, 32, 64\}\). We augment the data by preserving reverse-ordered sequences, yielding a 3-4\(\times\) increase in volume. All sequences are centroid-normalized to the origin. We then apply rigorous filtering to discard static or anomalous meshes, specifically removing sequences with inter-frame maximum vertex distance outside \([0.01, 1.0]\) or a face-to-vertex ratio exceeding 2.5. 

\noindent\textbf{Annotation and Statistics.} For text-motion alignment, we render the cleaned 4D meshes into video sequences and employ Qwen-2.5-VL~\cite{qwen25} to generate descriptive motion captions. \add{Regardless of source, every sequence is uniformly rendered with the same gradient-colored texture under fixed lighting (Fig.~\ref{fig:dymesh-xl}(e)) before captioning, so that AMASS/DT4D's untextured human motions and Objaverse-style textured object animations are treated through an identical, source-agnostic pipeline.}\rmark{R2-5} As detailed in Fig.~\ref{fig:dymesh-xl}, DyMesh-XL significantly surpasses the previous DyMesh~\cite{animateanymesh} dataset in scale, format support, and action diversity. The dataset is further partitioned into subsets based on maximum vertex counts (4K, 8K, 50K) to facilitate flexible training configurations, ultimately empowering AnimateAnyMesh++ to achieve state-of-the-art performance.

\noindent\add{\textbf{Licensing and Reproducibility.} The four sources aggregated into DyMesh-XL carry distinct licensing terms, which determine what we publicly release for each; a detailed per-source license breakdown and release policy is provided in the supplementary material (Sec.~II).}\rmark{R3-5}
\section{Experiments}
\label{sec:exp}
\noindent We evaluate AnimateAnyMesh++ against state-of-the-art methods, followed by comprehensive ablation studies validating our core architectural designs, and conclude with a discussion of limitations.

\subsection{Implementation Details}
\label{imp}
\noindent The DyMeshVAE-Flex architecture builds upon attention layers. In its encoder, the input sequence is segmented into non-overlapping 16-frame segments, maintaining temporal continuity by prepending the last 8 frames of the preceding segment, and utilizes positional encoding with 8 and 10 frequency components for vertices and relative trajectories, respectively. The encoder employs two \replace{$L=4$}{$L_{hop}=4$}\rmark{R3-m1} PLTA layers for local geometry encoding, followed by Farthest Point Sampling (1/8 ratio) and a \replace{Sync attention}{Sync Attention}\rmark{R3-m3} layer for downsampling and global aggregation. The decoder consists of 8 Sync Attention layers. All attention computations use a hidden dimension of 256, and the intermediate latent dimensions are compressed to 32 for both vertex and trajectory features. The SGTT model (Fig.~\ref{fig:sgtt}) is composed of 12 transformer blocks, with attention dimensions projected to 512. During training, an efficient batching strategy is used, involving zero-padding vertices and padding face indices with invalid values (up to 2.5 times the maximum vertex count). Inference is performed using rectified flow sampling over 64 uniformly sampled timesteps. The DyMeshVAE-Flex model is trained sequentially on the DyMesh-XL subsets (16, 32, and 64 frames, max 8192 vertices) for 30,000 iterations per subset with a batchsize of 16 and a  learning rate of 2e-4 using 32 H20 GPUs, taking approximately 72 hours. Subsequently, the SGTT model was trained sequentially on the same data for 50,000 iterations per subset (batchsize=8, learning rate=2e-4) on 32 H20 GPUs, requiring about 120 hours.

\subsection{Datasets}
\label{datasets}
\noindent For the training of our proposed DyMeshVAE-Flex network, we curate datasets by filtering the DyMesh-XL dataset for dynamic mesh sequences with fewer than 8,192 vertices. These sequences are subsequently segmented into 16, 32, and 64-frame clips, resulting in subsets containing 2.5M, 1.2M, and 415K samples, respectively. From these subsets, we meticulously select 512 examples for the validation set and another 512 for the test set. The selection criteria ensure that the chosen examples feature significant motion, contain no consecutive static frames, and have minimal subject overlap to promote diversity. Throughout this section, we denote these DyMesh-XL subsets as $\mathrm{DM\_<1>v\_<2>f\_<3>}$, where $\_<1>$ represents the maximum vertex count, $\_<2>$ indicates the number of frames, and $\_<3>$ specifies the data split (i.e., train, val, or test). To further assess the model's performance on larger meshes, we evaluate its compression-reconstruction capabilities on an additional set of 256 examples for each sequence length (16, 32, and 64 frames), where meshes have vertex counts ranging from over 8,192 to under 50,000.

To benchmark the generative performance of the SGTT Model and the overall AnimateAnyMesh++ framework, we assemble a diverse test bed of 50 models from Sketchfab and Mixamo. This collection spans five distinct categories: bipeds (20), quadrupeds (10), flying creatures (5), aquatic creatures (5), and inanimate objects (10). Furthermore, we employ LLMs to generate 10 common driving prompts for each category, which are used to systematically evaluate the generation quality and versatility of AnimateAnyMesh++. Please refer to the supplementary materials for details of the animation prompts.

\subsection{Evaluation Metrics}
\label{metrics}
\noindent \textbf{Reconstruction Metrics.} We evaluate the geometric fidelity of DyMeshVAE-Flex using Average Vertex Error (AVE), defined as the mean L2 distance between reconstructed and ground-truth vertices across all timesteps. To quantify structural soundness and detect unnatural deformations (e.g., mesh adhesion), we introduce the Anomalous Edge Ratio (\(\rho_{abn}^{\tau}\)). It calculates the proportion of edges whose length ratio between the reconstructed and input meshes exceeds a threshold \(\tau\) or falls below \(1/\tau\). We employ \(\rho_{abn}^{2}\), \(\rho_{abn}^{5}\), and \(\rho_{abn}^{10}\) to quantify mild, moderate, and severe anomalies, respectively. The formal mathematical definition is provided in the Supplementary Material.

\noindent \textbf{Animation Metrics.} Following previous methods~\cite{animateanymesh,animate3d}, we utilize VBench~\cite{vbench} metrics to assess the rendered videos: I2V Subject Similarity (I2V) for geometric consistency, Motion Smoothness (M.sm), Aesthetic Quality (Aest.Q), and Overall Consistency (O.C) for text-motion alignment. To explicitly evaluate the inherent 4D mesh dynamics, we introduce Average Moving Distance (AMD) to measure the magnitude of generated motion via average inter-frame vertex displacement, alongside the aforementioned \(\rho_{abn}^{2}\) to assess local shape preservation and resistance to trajectory sticking.

\noindent\add{\textbf{Local Rigidity and Self-Intersection Metrics.} The metrics above still cannot directly reveal two failure modes specific to mesh animation: locally non-rigid surface distortion and mesh self-penetration. We therefore introduce two additional targeted metrics, computed against each mesh's own rest pose and requiring no ground-truth motion: the Local Rigidity Error (LRE), an ARAP-style~\cite{arap} residual quantifying how far each vertex's local deformation deviates from a best-fit rigid (rotation-only) transformation, and the Self-Intersection Rate (SIR), the fraction of non-adjacent triangle pairs that newly self-intersect relative to the rest pose. Both metrics apply only to methods with an explicit, topology-consistent vertex trajectory (Animate3D, AnimateAnyMesh, AnimateAnyMesh++); formal definitions are provided in the Supplementary Material, and results and discussion in Table~\ref{tab:quant}.}\rmark{R3-4}

\subsection{Comparison with Previous Works}
\begin{figure*}[t]
  \centering
   \includegraphics[width=1.0\linewidth]{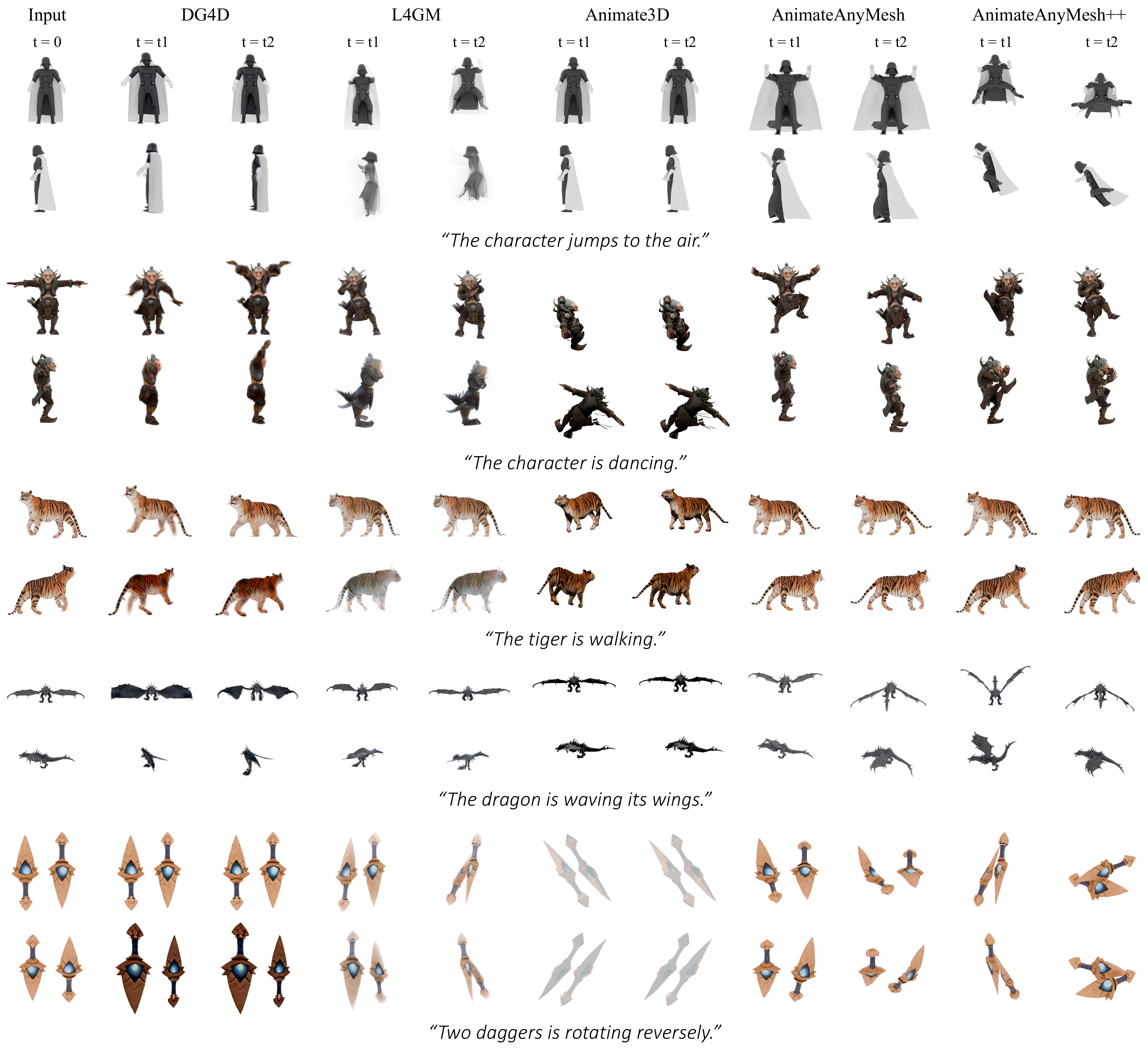}
   \caption{Qualitative comparison of text-to-4D generation. Given an initial 3D mesh (Input at $t=0$) and a driving text prompt, we compare our method (AnimateAnyMesh++) against state-of-the-art baselines. For each subject, we visualize the generated dynamics at two different timesteps ($t_1$ and $t_2$) across two distinct viewpoints (top and bottom rows per example). \add{Full temporal video renders of these examples for every compared method are provided in the supplementary video.}\rmark{R2-3}}
   \label{fig:comp}
\end{figure*}

\noindent To comprehensively evaluate the 4D generation performance of our method, we compare it against four representative state-of-the-art approaches: DreamGaussian4D (DG4D)~\cite{dreamgaussian4d}, L4GM~\cite{l4gm}, Animate3D~\cite{animate3d}, and our conference predecessor, AnimateAnyMesh~\cite{animateanymesh}. These baselines are carefully selected to cover a diverse range of methodologies, including Gaussian-based (DG4D, L4GM) and mesh-based (Animate3D, AnimateAnyMesh) 4D generation, as well as optimization-based (DG4D, Animate3D) and feed-forward (L4GM, AnimateAnyMesh) paradigms. \replace{Note that for L4GM, the reference videos are generated using the state-of-the-art video generator Wan2.2.}{Note that L4GM requires a monocular driving video as input. Rather than synthesizing this video with a general-purpose image-to-video generator (e.g., Wan2.2~\cite{wan}), which we found prone to visible artifacts on this background-less, square, purely-rendered visual style, we instead use our own method's front-view rendered animation (i.e., the front-view render produced by AnimateAnyMesh++ for the same test case) as the driving video for L4GM. This isolates L4GM's own novel-view reconstruction quality from confounds introduced by driving-video artifacts.}\rmark{R1-6} Furthermore, since the specialized 4D metrics introduced in Sec.~\ref{metrics} are specifically designed to evaluate mesh animations, we only compute and report these measurements for the mesh-based approaches, namely Animate3D, AnimateAnyMesh, and our AnimateAnyMesh++.

\noindent\textbf{Qualitative Comparison.} 
As illustrated in Fig.~\ref{fig:comp}, DG4D relies on Stable Video Diffusion (SVD)~\cite{svd}—a relatively early video generation model—as its driving signal generator, which restricts the generated animations to very subtle motion amplitudes. Furthermore, it exhibits poor spatio-temporal consistency and degraded local geometry and texture quality when rendered from novel views. L4GM, which feed-forwardly generates 4D Gaussians directly from monocular videos, achieves high visual quality in the reference view. However, it struggles with unreasonable geometric distortions and temporal inconsistencies under novel views (e.g., the 1st and 4th examples in Fig.~\ref{fig:comp}).
Animate3D adopts a pipeline that first generates orthogonal-view dynamic videos of the target object and subsequently solves for mesh vertex trajectories via optimization. Although it performs well in certain cases, it generally produces limited motion dynamics across most scenarios (e.g., the 1st, 2nd, 4th, and 5th examples). More critically, its optimization process requires per-instance hyperparameter tuning, significantly hindering its practical applicability. Our conference version, AnimateAnyMesh, pioneered the feed-forward vertex trajectory prediction paradigm for mesh animation, enabling fast and convenient animation of arbitrary meshes. Nevertheless, constrained by its fixed-length compression strategy and limited training data, the diversity and accuracy of its generated motions remain suboptimal.
In contrast, by leveraging the newly proposed large-scale, high-quality 4D dataset (DyMesh-XL) and the more robust dynamic mesh vertex trajectory compression network (DyMeshVAE-Flex), our AnimateAnyMesh++ successfully overcomes these limitations. As demonstrated in Fig.~\ref{fig:comp}, our method significantly outperforms all existing approaches in terms of motion dynamics, semantic alignment, and generation diversity.

\noindent\textbf{Quantitative Comparison.} 
As summarized in Tab.~\ref{tab:quant}, we quantitatively evaluate the generated results using VBench~\cite{vbench} for video-level metrics and specialized 4D metrics for structural integrity and motion dynamics. 
In terms of video metrics, our AnimateAnyMesh++ consistently achieves the highest scores across foreground consistency (I2V), motion smoothness (M.Sm), aesthetic quality (Aest.Q), and text alignment (O.C). 
The advantages of our method are even more pronounced in the 4D metrics, which directly measure the quality of the underlying dynamic meshes. Our method achieves the highest dynamic motion magnitude (AMD of 0.0043), quantitatively verifying our qualitative observation that AnimateAnyMesh++ can generate pronounced, diverse, and large-amplitude motions, effectively overcoming the subtle motion limitations of Animate3D (0.0011). More importantly, despite generating significantly larger motions, our method achieves the lowest Anomalous Edge Ratio (\(\rho_{abn}^{2}\) of 0.0050). This indicates a strong capability for local shape preservation and exceptional resistance to trajectory sticking. This substantial improvement over our conference predecessor, AnimateAnyMesh (0.0089), can be attributed to the high-quality DyMesh-XL dataset and the robust representation power of the DyMeshVAE-Flex, which collaboratively ensure strict geometric consistency even under complex and large-scale articulations.
\add{We further validate geometric quality with the Local Rigidity Error (LRE) and Self-Intersection Rate (SIR) introduced in Sec.~\ref{metrics}, also reported in Tab.~\ref{tab:quant} for the three methods with an explicit vertex trajectory. Although Animate3D attains the lowest LRE/SIR, this mainly reflects its near-static, low-AMD failure mode (Fig.~\ref{fig:comp}) rather than higher quality, since a barely-moving mesh is trivially both rigid and non-self-intersecting. Against our own conference predecessor, however, AnimateAnyMesh++ simultaneously achieves a larger motion magnitude (AMD $0.0043>0.0035$), a lower LRE ($2.7\times10^{-5}<3.0\times10^{-5}$), and a lower SIR ($0.37\%<0.54\%$), confirming that its larger motion amplitude is not achieved at the expense of geometric quality.}\rmark{R3-4}
Finally, we evaluate the inference efficiency. Compared to optimization-based paradigms like DG4D (\(\sim\)10 mins) and Animate3D (\(\sim\)14 mins), our feed-forward approach drastically reduces the generation time to mere seconds. Unlike our conference version, which necessitates training separate models for specific sequence lengths (e.g., distinct models for 16 and 32 frames), AnimateAnyMesh++ elegantly supports variable-length inference within a single unified model. Remarkably, scaling the generation from 16 frames (7.3s) to 32 frames (9.5s) and even 64 frames (13.5s) does not incur a prohibitive increase in computational overhead. This high efficiency is achieved through the architectural design of alternating spatio-temporal attention mechanisms and the parallel decoding strategy of our VAE, which seamlessly accommodate an increasing number of temporal chunks without significantly compromising the inference speed.

\begin{table*}[t!]
  \centering
  \add{\caption{Comprehensive comparison with state-of-the-art methods, consolidating video-level metrics, 4D-specific mesh metrics, user-study results (Sec.~\ref{sec:exp}-D), and inference time into a single table. Except for inference time, all metrics for AnimateAnyMesh and our AnimateAnyMesh++ are evaluated on 16-frame generation results; inference time is measured on a single NVIDIA A800 GPU. User Study columns report the percentage of votes received by each method along each evaluation axis (out of 500 votes per axis, i.e., 50 participants \(\times\) 10 test cases).\label{tab:quant}}}

  \resizebox{\textwidth}{!}{  
    \begin{tabular}{l|cccc|cccc|ccc|c}
      \toprule
      \multirow{2}{*}{\textbf{Method}}                  & \multicolumn{4}{c|}{\textbf{Video Metrics}}                                 & \multicolumn{4}{c|}{\textbf{4D Metrics}}          & \multicolumn{3}{c|}{\add{\textbf{User Study (\% votes)}}}          & \multirow{2}{*}{\textbf{Time}~\(\downarrow\)} \\ \cline{2-12}
                                                        & \textbf{I2V}~\(\uparrow\) & \textbf{M.Sm}~\(\uparrow\) & \textbf{Aest.Q}~\(\uparrow\) & \textbf{O.C}~\(\uparrow\) & \textbf{AMD}~\(\uparrow\) & \textbf{\(\boldsymbol{\rho_{abn}^{2}}\)}~\(\downarrow\) & \add{\textbf{LRE}~\(\downarrow\)} & \add{\textbf{SIR}~\(\downarrow\)} & \add{\textbf{Shape Pres.}~\(\uparrow\)} & \add{\textbf{Text Align.}~\(\uparrow\)} & \add{\textbf{Motion Qual.}~\(\uparrow\)} &                                         \\
    \midrule
DG4D~\cite{dreamgaussian4d} & 0.811               & 0.926                & 0.476                  & 0.204                   & /                   & /                   & \add{/}                   & \add{/}                   & \add{2.4\%}                   & \add{5.0\%}                   & \add{4.8\%}                   & \(\sim\)10min (14f)                                                 \\
L4GM~\cite{l4gm}            & 0.844               & 0.992                & 0.464                  & 0.181                   & /                   & /                   & \add{/}                   & \add{/}                   & \add{6.6\%}                   & \add{12.0\%}                  & \add{11.4\%}                  & \(\sim\)30s (32f)                                                   \\
\add{4Diffusion\(^{\dagger}\)~\cite{4diffusion}} & \add{/} & \add{/} & \add{/} & \add{/} & \add{/} & \add{/} & \add{/} & \add{/} & \add{3.6\%} & \add{11.2\%} & \add{10.8\%} & \add{/} \\
Animate3D~\cite{animate3d}  & 0.936               & 0.992                & 0.526                  & 0.201                   & 0.0011                   & 0.0067                   & \add{\(2.5\times10^{-5}\)}                   & \add{0.29\%}                   & \add{21.6\%}                   & \add{6.0\%}                   & \add{6.6\%}                   & \(\sim\)14min (16f)                                                \\
\midrule
AnimateAnyMesh~\cite{animateanymesh}              & 0.954               & \textbf{0.995}                & 0.539                  & 0.219                   & 0.0035                   & 0.0089                   & \add{\(3.0\times10^{-5}\)}                   & \add{0.54\%}                   & \add{28.0\%}                   & \add{27.8\%}                   & \add{26.8\%}                   & \textbf{\makecell{6.4s (16f) \\ 6.8s (32f)}}            \\
\midrule
\textbf{AnimateAnyMesh++ (Ours)}                  & \textbf{0.956}      & \textbf{0.995}       & \textbf{0.551}         & \textbf{0.227}          & \textbf{0.0043}          & \textbf{0.0050}          & \add{\(\mathbf{2.7\times10^{-5}}\)}          & \add{\textbf{0.37\%}}          & \add{\textbf{37.8\%}}          & \add{\textbf{38.0\%}}          & \add{\textbf{39.6\%}}          & \makecell{7.3s (16f) \\ 9.5s (32f) \\ 13.5s (64f)}   \\
      \bottomrule
    \end{tabular}
  } 
  \\[2pt]
  \add{\footnotesize \(^{\dagger}\)4Diffusion is evaluated only via the User Study (Sec.~\ref{sec:exp}-D); its officially released checkpoint outputs only 8 frames per view, precluding a fair automated-metric comparison against the 16-frame baselines used for every other column (see qualitative results in the supplementary video). ``/'' for AMD/\(\rho_{abn}^{2}\)/LRE/SIR indicates methods without an explicit, topology-consistent vertex trajectory (Gaussian-splat-based DG4D/L4GM, and 4Diffusion's multi-view-video-only output), to which these metrics do not apply.}
\end{table*}

\add{\noindent\textbf{Variable-Length Generation Analysis.} Since variable-length generation is one of our core contributions, we further evaluate whether generation quality is stable across sequence lengths, rather than only at the single 16-frame length used in Tab.~\ref{tab:quant}. As shown in Tab.~\ref{tab:varlen}, we report the full metric suite for AnimateAnyMesh++ at 16, 32, 48, and 64 frames, generated by the same unified model; this is an ablation-style self-comparison rather than a cross-method comparison, since most baselines in Tab.~\ref{tab:quant} do not support an analogous variable-length setting. Notably, 48 frames is never seen during training (Sec.~\ref{datasets}), so its inclusion directly probes generalization to an unseen sequence length. All metrics remain essentially stable across the four lengths, with only a mild, expected drift in O.C and \(\rho_{abn}^{2}\) as sequences get longer, and performance at the untrained 48-frame length is statistically indistinguishable from the trained lengths -- direct evidence that AnimateAnyMesh++ genuinely generalizes across sequence lengths rather than merely memorizing the ones seen during training.}

\begin{table}[t]
  \centering
  \add{\caption{Evaluation of AnimateAnyMesh++ across different generation lengths. Only 16-, 32-, and 64-frame sequences are used during training (Sec.~\ref{datasets}); 48 frames is never seen in training and directly probes generalization to an unseen, interpolated sequence length. All metrics are computed on the same test bed as Tab.~\ref{tab:quant}.\label{tab:varlen}}}
  \resizebox{\columnwidth}{!}{%
  \add{\begin{tabular}{c|cccc|cc}
    \toprule
    \textbf{\#Frames} & \textbf{I2V}~\(\uparrow\) & \textbf{M.Sm}~\(\uparrow\) & \textbf{Aest.Q}~\(\uparrow\) & \textbf{O.C}~\(\uparrow\) & \textbf{AMD}~\(\uparrow\) & \textbf{\(\boldsymbol{\rho_{abn}^{2}}\)}~\(\downarrow\) \\
    \midrule
    16             & 0.956 & 0.995 & 0.551 & 0.227 & 0.0043 & 0.0050 \\
    32             & 0.956 & 0.994 & 0.549 & 0.224 & 0.0046 & 0.0052 \\
    48\(^{\dagger}\) & 0.955 & 0.994 & 0.550 & 0.223 & 0.0045 & 0.0057 \\
    64             & 0.956 & 0.995 & 0.550 & 0.220 & 0.0045 & 0.0063 \\
    \bottomrule
  \end{tabular}}
  }
  \\[2pt]
  \add{\footnotesize \(^{\dagger}\)48-frame sequences are not included in any training subset.}\rmark{R3-3}
\end{table}


\add{\noindent\textbf{User Study.} Since automated metrics may not fully capture perceived quality, we further conduct a user study to complement Tab.~\ref{tab:quant}. We recruit 50 participants and randomly select 10 test cases from our benchmark; for each case, every method under comparison (the five methods evaluated quantitatively above, plus 4Diffusion~\cite{4diffusion} as an additional multi-view-video baseline whose short, 8-frame output length precludes automated-metric comparison, see Tab.~\ref{tab:quant}) generates a result, and participants are asked to pick the single best-performing method along three independent axes: shape preservation, text alignment, and motion quality. The rightmost three columns of Tab.~\ref{tab:quant} report the percentage of votes (out of 500 total votes per axis, i.e., 50 participants \(\times\) 10 cases) received by each method. AnimateAnyMesh++ receives the plurality of votes on all three axes (37.8\%, 38.0\%, and 39.6\%, respectively), consistently and substantially ahead of every baseline, including our conference version AnimateAnyMesh (28.0\%, 27.8\%, and 26.8\%). This human-perceptual evidence corroborates the automated-metric improvements reported in Tab.~\ref{tab:quant} and directly addresses the concern that VBench-style metrics alone may not reliably reflect user-perceived quality.}\rmark{R2-4}

\subsection{Ablation Studies and Analysis}
\noindent In this subsection, we conduct a series of comprehensive ablation studies to systematically dissect our proposed DyMeshVAE-Flex and validate the efficacy of its key components. Our primary architectural and methodological innovations include: (1) a novel chunk-based temporal compression mechanism that extends sequence length capabilities; (2) the integration of vertex normal information to enhance surface detail; (3) the proposed Power-Law Topology-Aware (PLTA) attention mechanism designed to mitigate adhesion artifacts; and (4) the Time-Dependent Gradient Weighting (TDGW) strategy for ensuring seamless and smooth transitions between temporal chunks.

To isolate the contribution of each component, we establish a baseline model and incrementally add our proposed modules, evaluating the impact on reconstruction quality and artifact suppression at each step. Specifically, we will demonstrate how the chunk-based mechanism maintains high fidelity over long sequences, how vertex normals and PLTA attention collaboratively improve reconstruction accuracy and resolve mesh adhesion issues, and finally, how the TDGW strategy effectively eliminates temporal discontinuities. Both quantitative metrics and qualitative visualizations are presented to provide a thorough analysis.

\begin{figure}[t]
  \centering
   \includegraphics[width=1.0\linewidth]{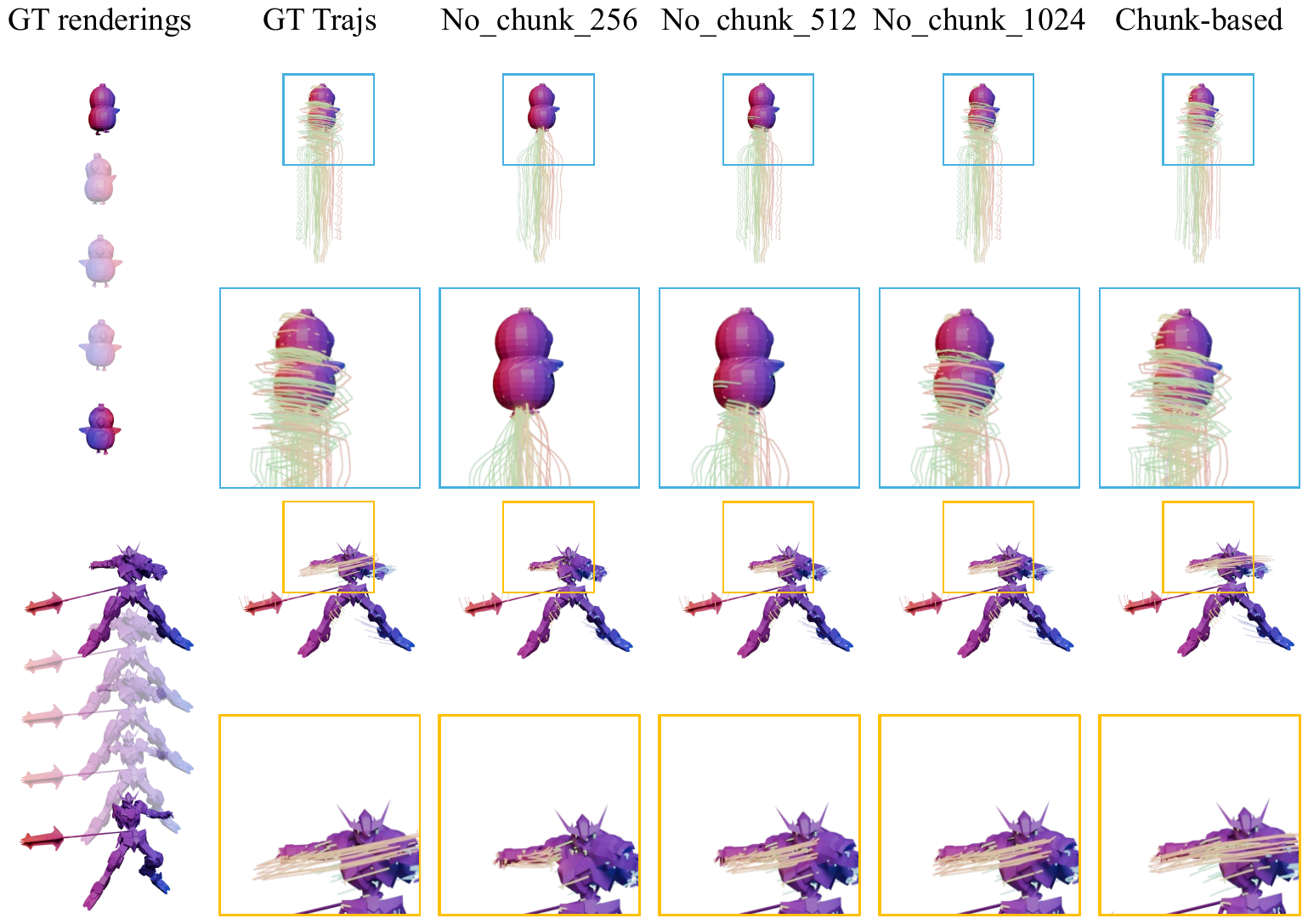}
   \caption{Ablation study on our proposed chunk-based trajectory compression against a holistic (no-chunk) baseline. From left to right: ground truth (GT) renderings, GT vertex trajectories, holistic compression with 256, 512, and 1024 channels, and our chunk-based method. We visualize the trajectories of vertices as colored lines.}
   \label{fig:abl_chunk}
\end{figure}

\noindent \textbf{Chunk-based Temporal Compression.} 
In the original DyMeshVAE~\cite{animateanymesh}, the information of each trajectory is compressed into the channel dimension. This approach introduces a significant problem: as the sequence length increases, one must either incur prohibitive computational costs by using a very large feature dimension for attention calculations and intermediate feature storage, or suffer a substantial degradation in reconstruction quality by keeping the feature dimension constant.

In contrast, our proposed DyMeshVAE-Flex adopts a chunk-based strategy, where each sequence is partitioned into several chunks of a fixed length ($T=16$). This constant-size partitioning offers excellent scalability, effectively decoupling the latent feature dimension from the number of chunks. Furthermore, due to our Sync Attention design, where trajectory features are simply mapped using attention maps derived from vertex features, the increase in computational cost is marginal as the number of chunks grows, laying a solid foundation for long-sequence dynamic mesh compression-reconstruction. Fig.~\ref{fig:abl_chunk} and Tab.~\ref{tab:chunk} presents the comparison of reconstruction quality and inference time between the original holistic-sequence compression method (with varying channel dimensions) and our chunk-based approach on 64-frame sequences.

\begin{table}[h!]
\centering
\caption{
    Ablation study on the proposed ``Chunk-based'' strategy. All experiments are conducted on 64-frame sequences. The best results are shown in \textbf{bold}.
}
\begin{tabular}{lcccccc}
\toprule
\textbf{Experiment} & \textbf{Dim} & \textbf{AVE} ($\downarrow$) & \textbf{$\boldsymbol{\rho_{abn}^{2}}$} ($\downarrow$) & \textbf{$\boldsymbol{\rho_{abn}^{5}}$} ($\downarrow$) & \textbf{Time} ($\downarrow$) \\
\midrule
\multirow{3}{*}{No Chunk} & 256  & 0.0126 & 0.0080 & 0.0023 & $\mathbf{\times 1.00}$  \\
                          & 512  & 0.0102 & 0.0069 & 0.0019 & $\times 1.98$  \\
                          & 1024 & 0.0083 & \textbf{0.0060} & \textbf{0.0014} & $\times 4.03$  \\
\midrule
Chunk-based               & 256  & \textbf{0.0070} & 0.0076 & 0.0021 & $\times 1.04$ \\
\bottomrule
\end{tabular}
\label{tab:chunk}
\end{table}

As shown in Fig.~\ref{fig:abl_chunk}, the baseline methods using holistic trajectory compression (No\_chunk\_256, No\_chunk\_512) fail to achieve high-fidelity reconstruction for motions characterized by high speeds and large ranges. Even when using 1,024 channels for trajectory feature representation and computation, a significant gap in reconstruction quality still exists compared to our chunk-based approach.
Similarly, Tab.~\ref{tab:chunk} provides quantitative evidence that reinforces this finding. The proposed chunk-based strategy achieves a significant improvement in reconstruction accuracy over the holistic-sequence compression approach, all while incurring only a marginal increase in computational cost with the same number of channels. Furthermore, when compared to the full-sequence baselines employing higher feature dimensions, our chunk-based approach not only demonstrates superior reconstruction quality but also surpasses them in terms of inference time, showcasing its excellent overall performance.

\noindent \textbf{Vertex Normal Injection.} 
Vertex normals provide rich local geometric information, and their effective integration can both improve reconstruction quality and significantly reduce adhesion artifacts. The encoding method, however, is critical to performance. We evaluate three main approaches: concatenation with coordinates prior to positional encoding (Normal Enc1), concatenation with vertex features before neighborhood encoding (Normal Enc2), and concatenation after neighborhood encoding (Normal Enc3). Tab.~\ref{tab:normal_pe} presents a quantitative comparison of these strategies against a baseline without normal information.

\begin{table}[h!]
\centering
\caption{
    Ablation study on the vertex normal injection strategy. All experiments are conducted on 64-frame sequences. The best results are shown in \textbf{bold}.
}
\begin{tabular}{lccccc}
\toprule
\textbf{Experiment} &  \textbf{AVE} ($\downarrow$) & \textbf{$\boldsymbol{\rho_{abn}^{2}}$} ($\downarrow$) & \textbf{$\boldsymbol{\rho_{abn}^{5}}$} ($\downarrow$) & \textbf{$\boldsymbol{\rho_{abn}^{10}}$} ($\downarrow$) \\
\midrule
No Normal   & 0.0126 & 0.0080 & 0.0023 & 0.0012 \\
Normal~Enc1   & 0.0158 & 0.0123 & 0.0030 & 0.0017 \\
Normal~Enc2   & \textbf{0.0115} & \textbf{0.0051} & \textbf{0.0010} & \textbf{0.0006} \\
Normal~Enc3   & 0.0118 & 0.0059 & 0.0013 & 0.0007 \\
\bottomrule
\end{tabular}
\label{tab:normal_pe}
\end{table}

\begin{figure}[t]
  \centering
   \includegraphics[width=1.0\linewidth]{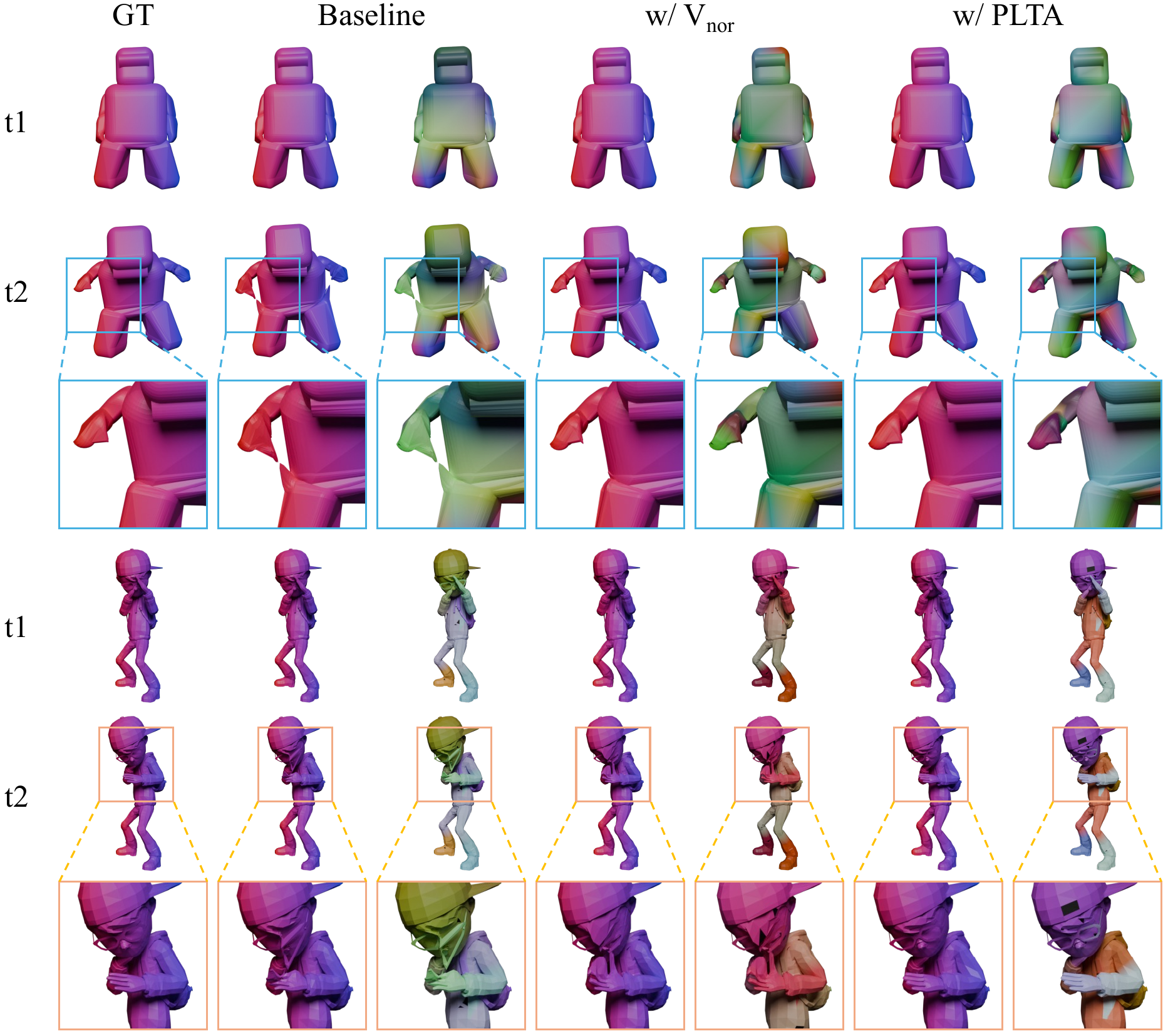}
   \caption{Ablation study on the contributions of Vertex Normal Injection ($V_{nor}$) and the proposed Power-Law Topology-Aware (PLTA) attention module. We visualize the UMAP projection of the final vertex features onto the mesh surface at two different timestamps (t1, t2) to illustrate the distributions of the  encoded query features.}
   \label{fig:abl_stuck}
\end{figure}

As shown in Tab.~\ref{tab:normal_pe}, both Normal Enc2 and Normal Enc3 demonstrate significant improvements over the baseline (which omits normal information) in terms of reconstruction quality and the mitigation of adhesion artifacts, with Normal Enc2 achieving the best performance. The rationale is that, with the Normal Enc2 strategy, the vertex normal information is aggregated via the neighborhood matrix alongside other features, thereby enhancing the model's perception of the local shape. Besides, it is evidenced in Fig.~\ref{fig:abl_stuck} that the injection of vertex normal information helps eliminate the trajectory adhesion issue, which brings significant visual improvements. 

Conversely, Normal Enc1 leads to a performance degradation. This is because positional encoding is designed to project the continuous coordinate space into a representation with superimposed high-frequency components to enhance local discriminability. However, this process is detrimental to normal vectors, for which continuity itself contains rich semantic information. Disrupting this continuity interferes with the reconstruction process. For instance, two adjacent points on a flat plane have nearly identical normals, representing a smooth surface. Projecting them using periodic high-frequency functions breaks this smoothness, violating the strong prior that neighboring points on a plane should exhibit highly correlated motion. This, in turn, degrades the reconstruction quality to some extent.

\noindent \textbf{Power-Law Topology-Aware (PLTA) Attention Mechanism.} 
As one of the core designs of DyMeshVAE-Flex, our Power-Law Topology-Aware (PLTA) attention significantly expands the receptive field of each vertex. It achieves this by aggregating features from an exponentially growing neighborhood, defined by matrix powers of the adjacency matrix, and weighting them with a fixed decay coefficient. This design effectively prevents the feature blurring issue often caused by using uniform weights for all hops.

Here, we conduct a quantitative analysis of two key hyperparameters in PLTA attention: the number of hop bands, \replace{denoted as nhops}{$L_{hop}$}\rmark{R3-m1}, and the hop decay coefficient, \replace{$\alpha$-hops}{$\gamma_{hop}$}\rmark{R3-m1}. For the hop decay coefficient, we also include a learnable variant (initialized to all ones) for comparison. The results are presented in Tab.~\ref{tab:plta}.

\begin{table}[h!]
\centering
\caption{
    Ablation studies on the key hyperparameters of our PLTA attention module: the number of hops (\replace{nhops}{$L_{hop}$}) and the mixing weight (\replace{$\alpha$-hops}{$\gamma_{hop}$}). We first determine the optimal $L_{hop}$ and then use it to find the best $\gamma_{hop}$. All experiments are conducted on 64-frame sequences. The best results in each group are shown in \textbf{bold}.
}
\label{tab:plta}
\begin{tabular}{lccccc}
\toprule
\textbf{Param} & \textbf{Value} & \textbf{AVE} ($\downarrow$) & \textbf{$\boldsymbol{\rho_{abn}^{2}}$} ($\downarrow$) & \textbf{$\boldsymbol{\rho_{abn}^{5}}$} ($\downarrow$) & \textbf{$\boldsymbol{\rho_{abn}^{10}}$} ($\downarrow$) \\
\midrule
\multirow{8}{*}{\replace{nhops}{$L_{hop}$}} 
 & 0 & 0.0287 & 0.0293 & 0.0080 & 0.0031 \\
 & 1 & 0.0126 & 0.0080 & 0.0023 & 0.0012 \\
 & 2 & 0.0115 & 0.0066 & 0.0014 & 0.0006 \\
 & 3 & 0.0105 & 0.0052 & 0.0010 & 0.0005 \\
 & 4 & \textbf{0.0102} & \textbf{0.0047} & 0.0008 & \textbf{0.0004} \\ 
 & 5 & 0.0104 & \textbf{0.0047} & \textbf{0.0007} & \textbf{0.0004} \\
\midrule 
\multirow{5}{*}{\replace{$\alpha$-hops}{$\gamma_{hop}$}} 
 & 0.3 & 0.0100 & 0.0049 & 0.0010 & 0.0005 \\
 & 0.5 & \textbf{0.0102} & \textbf{0.0047} & 0.0008 & \textbf{0.0004}  \\
 & 0.7 & 0.0111 & 0.0047 & \textbf{0.0007} & 0.0004 \\
 & 1.0 & 0.0120 & 0.0060 & 0.0010 & 0.0006 \\
 & learn & 0.0115 & 0.0048 & 0.0008 & 0.0005 \\
\bottomrule
\end{tabular}
\end{table}

\begin{table*}[t!]
\centering
\caption{Comprehensive ablation study of our proposed components. The table is divided into three parts: (I) the individual impact of each component on the baseline; (II) the cumulative effect of progressively adding components; (III) our final model with all components enabled. $\downarrow$ indicates lower is better, while $\uparrow$ indicates higher is better.}
\label{tab:ablation_hybrid}
\resizebox{\textwidth}{!}{%
\begin{tabular}{l cccc cccc cccc}
\toprule
\multirow{2}{*}{\textbf{Experiment}} & \multicolumn{4}{c}{\textbf{Components}} & \multicolumn{4}{c}{\textbf{DM\_8192v\_64f\_test}} & \multicolumn{4}{c}{\textbf{DM\_50000v\_64f\_test}} \\
\cmidrule(lr){2-5} \cmidrule(lr){6-9} \cmidrule(lr){10-13}
& (a) Flex & (b) PLTA & (c) $V_{\text{nor}}$ & (d) TDGW & \textbf{AVE} $\downarrow$ & \textbf{$\boldsymbol{\rho_{abn}^{2}}$} ($\downarrow$) & \textbf{$\boldsymbol{\rho_{abn}^{5}}$} ($\downarrow$) & \textbf{$\boldsymbol{\rho_{abn}^{10}}$} ($\downarrow$) & \textbf{AVE} $\downarrow$ & \textbf{$\boldsymbol{\rho_{abn}^{2}}$} ($\downarrow$) & \textbf{$\boldsymbol{\rho_{abn}^{5}}$} ($\downarrow$) & \textbf{$\boldsymbol{\rho_{abn}^{10}}$} ($\downarrow$) \\
\midrule
Baseline & & & & & 0.0126 & 0.0080 & 0.0023 & 0.0012 & 0.0143 & 0.0175 & 0.0030 & 0.0013 \\
\midrule 
w/ (a) Flex only & \checkmark & & & & 0.0070 & 0.0076 & 0.0021 & 0.0010 & 0.0086 & 0.0169 & 0.0029 & 0.0012  \\
w/ (b) PLTA only & & \checkmark & & & 0.0102 & 0.0047 & 0.0008 & 0.0004 &  0.0122 & 0.0104 & 0.0016 & 0.0005 \\
w/ (c) $V_{\text{nor}}$ only & & & \checkmark & & 0.0115 & 0.0051 & 0.0010 & 0.0006 & 0.0130 & 0.0110 & 0.0018 & 0.0007 \\
w/ (a) Flex (d) TDGW   & \checkmark & & & \checkmark & 0.0066 & 0.0075 & 0.0020 & 0.0010 & 0.0082 & 0.0167 & 0.0028 & 0.0012 \\
\midrule 
M1: Baseline + (a) & \checkmark & & & & 0.0070 & 0.0076 & 0.0021 & 0.0010 & 0.0086 & 0.0169 & 0.0029 & 0.0012 \\
M2: M1 + (b) & \checkmark & \checkmark & & & 0.0063 & 0.0042 & 0.0006 & 0.0003 & 0.0079 & 0.0097 & 0.0015 & 0.0004 \\
M3: M2 + (c) & \checkmark & \checkmark & \checkmark & & 0.0054 & \textbf{0.0035} & \textbf{0.0005} & \textbf{0.0002} & 0.0070 & \textbf{0.0090} & \textbf{0.0014} & \textbf{0.0002} \\
\midrule 
\textbf{Ours (Full)} & \checkmark & \checkmark & \checkmark & \checkmark & \textbf{0.0047} & \textbf{0.0035} & \textbf{0.0005} & \textbf{0.0002} & \textbf{0.0063} & \textbf{0.0090} & \textbf{0.0014} & \textbf{0.0002} \\
\bottomrule
\end{tabular}
}
\end{table*}

From the table, we can observe that PLTA attention achieves its best performance when \replace{nhops}{$L_{hop}$} is set to 4 and \replace{alpha-hops}{$\gamma_{hop}$} is set to 0.5. Compared to both the baseline without neighborhood information (\replace{nhops=0}{$L_{hop}=0$}) and the original DyMeshVAE~\cite{animateanymesh} setting (\replace{nhops=1}{$L_{hop}=1$}), our optimal PLTA configuration yields substantial performance gains. This underscores the importance of expanding the local receptive field for improving reconstruction quality and reducing adhesion artifacts.\rmark{R3-m1}

Furthermore, when comparing the fixed decay coefficient to the learnable variant, the fixed coefficient achieves superior performance. This is because a fixed value like 0.5 explicitly encodes a strong prior that features from closer neighboring vertices are more important, which proves to be more effective than allowing the model to learn these weights from scratch.

To qualitatively demonstrate the efficacy of PLTA attention, we present two challenging 4D sequences prone to trajectory adhesion artifacts in Fig.~\ref{fig:abl_stuck}. The baseline method, lacking PLTA, exhibits severe adhesion in both examples. This issue is particularly pronounced in the second case, where the trajectory entanglement corrupts the facial geometry, leading to significant visual degradation. In stark contrast, our method, augmented with PLTA, faithfully reconstructs both sequences. To investigate the underlying mechanism, we visualize the UMAP~\cite{umap} projection of the encoded vertex features. The visualization reveals that, for the baseline, features of the character's left hand and face are highly similar, causing ambiguous queries and subsequent adhesion. Conversely, PLTA, by aggregating local neighborhood information, effectively disentangles these features. Consequently, the hand and face features become well-separated, and the hand's feature representation aligns more closely with that of the arm, which is consistent with natural motion correlations. This compellingly demonstrates that our PLTA attention resolves adhesion artifacts by learning more discriminative feature representations.

\begin{figure}[t]
  \centering
   \includegraphics[width=1.0\linewidth]{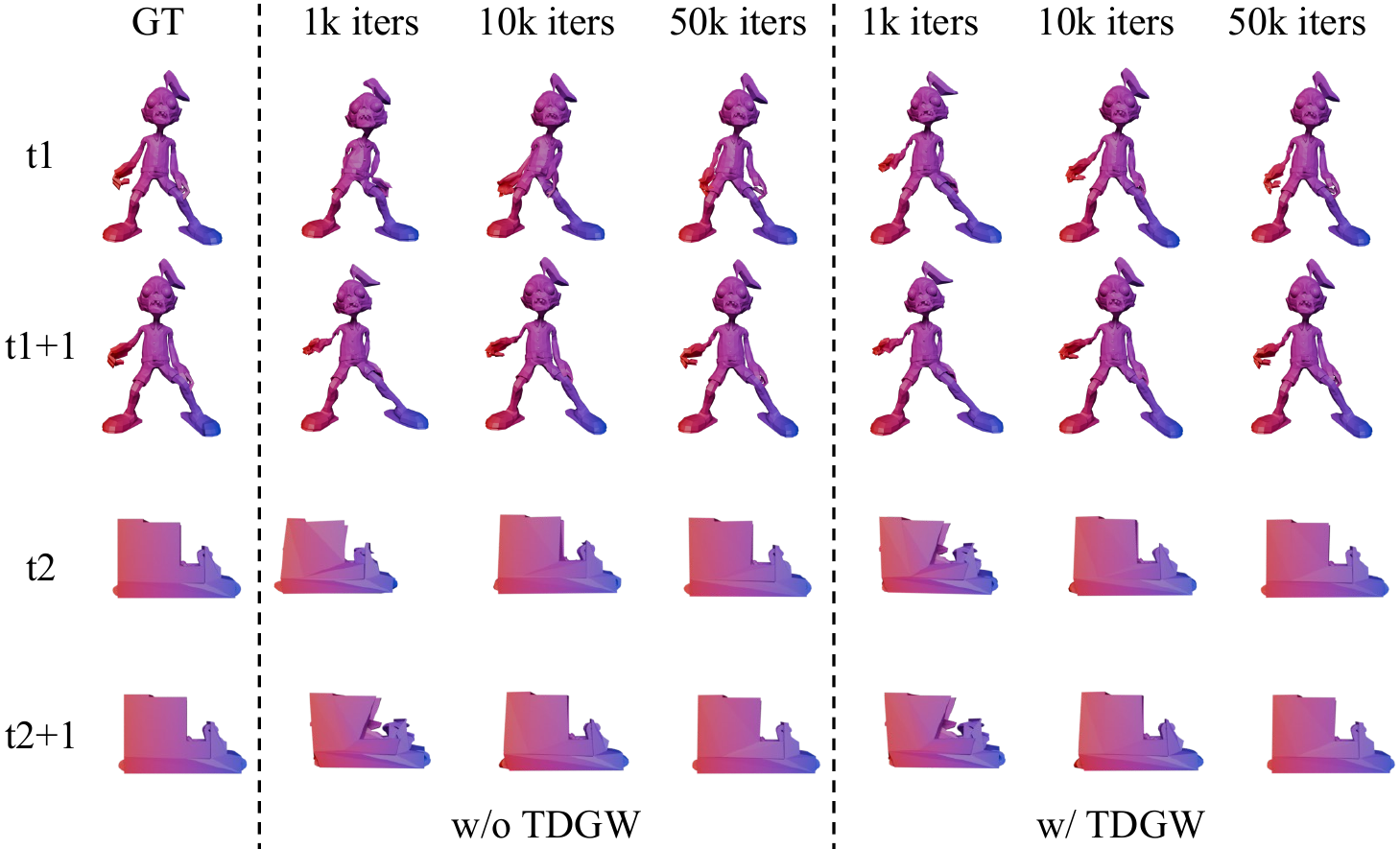}
   \caption{Ablation study demonstrating the effectiveness of our Time-Dependent Gradient Weighting (TDGW) module in resolving temporal discontinuities at chunk boundaries. We compare models with (w/ TDGW) and without (w/o TDGW) the module at different training stages. The rows t1 and t1+1 (t2 and t2+1) are specifically chosen to be the last frame of one chunk and the first frame of the subsequent chunk, respectively.}
   \label{fig:abl_tdgw}
\end{figure}

\noindent \textbf{Time-Dependent Gradient Weighting (TDGW) Strategy.} 
In the final stage of the DyMeshVAE-Flex decoder, the decoded chunks must be reassembled to form the complete trajectory. Without our proposed Time-Dependent Gradient Weighting (TDGW) strategy, this process results in noticeable discontinuities at the boundaries between chunks, as illustrated in Fig.~\ref{fig:abl_tdgw}. Although these jumps diminish as training progresses, they are never completely eliminated.

In contrast, as shown in Tab.~\ref{tab:tdgw}, our TDGW strategy enables the network to generate smooth and consistent trajectories from the very beginning. While introducing negligible computational overhead, it significantly enhances both the visual quality and the reconstruction accuracy. This lays a solid foundation for the generation of long 4D mesh sequences.

\begin{table}[h!]
\centering
\caption{
    Ablation study on the Time-Dependent Gradient Weighting (TDGW) strategy. All experiments are conducted on 64-frame sequences. The best results are shown in \textbf{bold}.
}
\begin{tabular}{lccccc}
\toprule
\textbf{Experiment} &  \textbf{AVE} ($\downarrow$) & \textbf{$\boldsymbol{\rho_{abn}^{2}}$} ($\downarrow$) & \textbf{$\boldsymbol{\rho_{abn}^{5}}$} ($\downarrow$) & \textbf{$\boldsymbol{\rho_{abn}^{10}}$} ($\downarrow$) \\
\midrule
w/o TDGW   & 0.0070 & 0.0076 & 0.0021 & \textbf{0.0010} \\
w/ TDGW    & \textbf{0.0066} & \textbf{0.0075} & \textbf{0.0020} & \textbf{0.0010} \\
\bottomrule
\end{tabular}
\label{tab:tdgw}
\end{table}

\noindent Building on the aforementioned contributions, DyMeshVAE-Flex achieves high-fidelity dynamic mesh reconstruction. To intuitively demonstrate the contribution of each component and their synergistic effects, we evaluated the reconstruction quality and trajectory adhesion metrics on the DM\_8192v\_64f\_test (512 samples) and DM\_50000v\_64f\_test (256 samples) datasets (check Sec.~\ref{datasets} for the details of testsets). As presented in Tab.~\ref{tab:ablation_hybrid}, each proposed component individually contributes to performance gains, and their combination allows DyMeshVAE-Flex to achieve exceptional performance in 4D mesh reconstruction.


\subsection{Limitation and Related Discussion}
\noindent Although AnimateAnyMesh++ pushes the performance ceiling of Text-Driven Universal Mesh Animation to new heights, it still has certain limitations. We summarize them into the following two points:

First, despite DyMesh-XL further expanding the scale of 4D mesh sequence data, its volume remains considerably limited compared to that of 3D or video datasets. This is primarily because 4D data is inherently scarce across the internet. DyMesh-XL has already incorporated nearly all freely and publicly available 4D data. However, referencing the standards set by video, image, and 3D generation, the current amount of 4D data is still insufficient to achieve truly high-quality driven generation.

Second, the annotation quality provided by current mainstream Vision-Language Models (VLMs) for videos of rendered dynamic objects is suboptimal. Even leading models such as GPT-4o, Gemini 2.5 Pro, and Qwen-VL 2.5 fail to produce highly accurate and detailed descriptions of the actions performed by rendered objects. This, in turn, restricts the text-alignment capability of the mesh animation model.
We leave these two points to be addressed in future work.
\section{Conclusion}
\noindent In this paper, we present AnimateAnyMesh++, a comprehensive framework that significantly advances the state of the art in text-driven universal mesh animation. Our approach is built upon several key innovations. At its core is DyMeshVAE-Flex, a novel autoencoder that achieves high-fidelity, variable-length sequence compression through a chunk-based strategy and our proposed Time-Dependent Gradient Weighting (TDGW). This design grants unprecedented flexibility, enabling the model to process sequences of arbitrary length during both training and inference.
To further enhance reconstruction quality, we introduced Power-Law Topology-Aware (PLTA) Attention and Vertex Normal Injection, which effectively expand the local receptive field to preserve fine-grained geometric details and ensure smooth, artifact-free motion. The success of our model is also supported by the introduction of the DyMesh-XL dataset, a large-scale collection of diverse and long-sequence 4D data that provides a robust foundation for training powerful generative models.
As a result, AnimateAnyMesh++ not only surpasses existing methods in both performance and efficiency but also uniquely enables the generation of ultra-long mesh animations, a capability that dramatically expands the practical application scope of this technology. We believe this work marks a significant step towards versatile and scalable 4D content creation and lays the groundwork for future explorations in this exciting domain.

\section*{Acknowledgements}
This work was supported by the NSFC (62225603).

\bibliographystyle{IEEEtran}
\bibliography{main}

@String(CVPR= {IEEE Conf. Comput. Vis. Pattern Recog.})

@String(ICCV= {Int. Conf. Comput. Vis.})

@String(TOG= {ACM Trans. Graph.})

@String(CVPR  = {CVPR})

@String(ICCV  = {ICCV})

@String(TOG   = {ACM TOG})

@inproceedings{clip,
  title={Learning transferable visual models from natural language supervision},
  author={Radford, Alec and Kim, Jong Wook and Hallacy, Chris and Ramesh, Aditya and Goh, Gabriel and Agarwal, Sandhini and Sastry, Girish and Askell, Amanda and Mishkin, Pamela and Clark, Jack and others},
  booktitle={International conference on machine learning},
  pages={8748--8763},
  year={2021},
  organization={PmLR}
}

@article{dreamfusion,
  title={Dreamfusion: Text-to-3d using 2d diffusion},
  author={Poole, Ben and Jain, Ajay and Barron, Jonathan T and Mildenhall, Ben},
  journal={arXiv preprint arXiv:2209.14988},
  year={2022}
}

@inproceedings{sd,
  title={High-resolution image synthesis with latent diffusion models},
  author={Rombach, Robin and Blattmann, Andreas and Lorenz, Dominik and Esser, Patrick and Ommer, Bj{\"o}rn},
  booktitle={Proceedings of the IEEE/CVF conference on computer vision and pattern recognition},
  pages={10684--10695},
  year={2022}
}

@inproceedings{sd3,
  title={Scaling rectified flow transformers for high-resolution image synthesis},
  author={Esser, Patrick and Kulal, Sumith and Blattmann, Andreas and Entezari, Rahim and M{\"u}ller, Jonas and Saini, Harry and Levi, Yam and Lorenz, Dominik and Sauer, Axel and Boesel, Frederic and others},
  booktitle={Forty-first international conference on machine learning},
  year={2024}
}

@article{cogvideox,
  title={Cogvideox: Text-to-video diffusion models with an expert transformer},
  author={Yang, Zhuoyi and Teng, Jiayan and Zheng, Wendi and Ding, Ming and Huang, Shiyu and Xu, Jiazheng and Yang, Yuanming and Hong, Wenyi and Zhang, Xiaohan and Feng, Guanyu and others},
  journal={arXiv preprint arXiv:2408.06072},
  year={2024}
}

@article{imagen,
  title={Photorealistic text-to-image diffusion models with deep language understanding},
  author={Saharia, Chitwan and Chan, William and Saxena, Saurabh and Li, Lala and Whang, Jay and Denton, Emily L and Ghasemipour, Kamyar and Gontijo Lopes, Raphael and Karagol Ayan, Burcu and Salimans, Tim and others},
  journal={Advances in neural information processing systems},
  volume={35},
  pages={36479--36494},
  year={2022}
}

@inproceedings{pointsto3d,
  title={Points-to-3d: Bridging the gap between sparse points and shape-controllable text-to-3d generation},
  author={Yu, Chaohui and Zhou, Qiang and Li, Jingliang and Zhang, Zhe and Wang, Zhibin and Wang, Fan},
  booktitle={Proceedings of the 31st ACM International Conference on Multimedia},
  pages={6841--6850},
  year={2023}
}

@inproceedings{sjc,
  title={Score jacobian chaining: Lifting pretrained 2d diffusion models for 3d generation},
  author={Wang, Haochen and Du, Xiaodan and Li, Jiahao and Yeh, Raymond A and Shakhnarovich, Greg},
  booktitle={Proceedings of the IEEE/CVF conference on computer vision and pattern recognition},
  pages={12619--12629},
  year={2023}
}

@article{prolificdreamer,
  title={Prolificdreamer: High-fidelity and diverse text-to-3d generation with variational score distillation},
  author={Wang, Zhengyi and Lu, Cheng and Wang, Yikai and Bao, Fan and Li, Chongxuan and Su, Hang and Zhu, Jun},
  journal={Advances in Neural Information Processing Systems},
  volume={36},
  pages={8406--8441},
  year={2023}
}

@inproceedings{zero123,
  title={Zero-1-to-3: Zero-shot one image to 3d object},
  author={Liu, Ruoshi and Wu, Rundi and Van Hoorick, Basile and Tokmakov, Pavel and Zakharov, Sergey and Vondrick, Carl},
  booktitle={Proceedings of the IEEE/CVF International Conference on Computer Vision},
  pages={9298--9309},
  year={2023}
}

@article{syncdreamer,
  title={Syncdreamer: Generating multiview-consistent images from a single-view image},
  author={Liu, Yuan and Lin, Cheng and Zeng, Zijiao and Long, Xiaoxiao and Liu, Lingjie and Komura, Taku and Wang, Wenping},
  journal={arXiv preprint arXiv:2309.03453},
  year={2023}
}

@inproceedings{objaverse,
  title={Objaverse: A universe of annotated 3d objects},
  author={Deitke, Matt and Schwenk, Dustin and Salvador, Jordi and Weihs, Luca and Michel, Oscar and VanderBilt, Eli and Schmidt, Ludwig and Ehsani, Kiana and Kembhavi, Aniruddha and Farhadi, Ali},
  booktitle={Proceedings of the IEEE/CVF Conference on Computer Vision and Pattern Recognition},
  pages={13142--13153},
  year={2023}
}

@article{objaverse-xl,
  title={Objaverse-xl: A universe of 10m+ 3d objects},
  author={Deitke, Matt and Liu, Ruoshi and Wallingford, Matthew and Ngo, Huong and Michel, Oscar and Kusupati, Aditya and Fan, Alan and Laforte, Christian and Voleti, Vikram and Gadre, Samir Yitzhak and others},
  journal={Advances in Neural Information Processing Systems},
  volume={36},
  year={2024}
}

@article{dreamcraft3d,
  title={Dreamcraft3d: Hierarchical 3d generation with bootstrapped diffusion prior},
  author={Sun, Jingxiang and Zhang, Bo and Shao, Ruizhi and Wang, Lizhen and Liu, Wen and Xie, Zhenda and Liu, Yebin},
  journal={arXiv preprint arXiv:2310.16818},
  year={2023}
}

@article{make-it-3d,
  title={Make-it-3d: High-fidelity 3d creation from a single image with diffusion prior},
  author={Tang, Junshu and Wang, Tengfei and Zhang, Bo and Zhang, Ting and Yi, Ran and Ma, Lizhuang and Chen, Dong},
  journal={arXiv preprint arXiv:2303.14184},
  year={2023}
}

@inproceedings{magic3d,
  title={Magic3d: High-resolution text-to-3d content creation},
  author={Lin, Chen-Hsuan and Gao, Jun and Tang, Luming and Takikawa, Towaki and Zeng, Xiaohui and Huang, Xun and Kreis, Karsten and Fidler, Sanja and Liu, Ming-Yu and Lin, Tsung-Yi},
  booktitle={Proceedings of the IEEE/CVF Conference on Computer Vision and Pattern Recognition},
  pages={300--309},
  year={2023}
}

@article{mav3d,
  title={Text-to-4d dynamic scene generation},
  author={Singer, Uriel and Sheynin, Shelly and Polyak, Adam and Ashual, Oron and Makarov, Iurii and Kokkinos, Filippos and Goyal, Naman and Vedaldi, Andrea and Parikh, Devi and Johnson, Justin and others},
  journal={arXiv preprint arXiv:2301.11280},
  year={2023}
}

@article{vae,
  title={Auto-encoding variational bayes},
  author={Kingma, Diederik P and Welling, Max},
  journal={arXiv preprint arXiv:1312.6114},
  year={2013}
}

@article{mvdream,
  title={Mvdream: Multi-view diffusion for 3d generation},
  author={Shi, Yichun and Wang, Peng and Ye, Jianglong and Long, Mai and Li, Kejie and Yang, Xiao},
  journal={arXiv preprint arXiv:2308.16512},
  year={2023}
}

@article{imagedream,
  title={ImageDream: Image-Prompt Multi-view Diffusion for 3D Generation},
  author={Wang, Peng and Shi, Yichun},
  journal={arXiv preprint arXiv:2312.02201},
  year={2023}
}

@article{instant3d,
  title={Instant3d: Fast text-to-3d with sparse-view generation and large reconstruction model},
  author={Li, Jiahao and Tan, Hao and Zhang, Kai and Xu, Zexiang and Luan, Fujun and Xu, Yinghao and Hong, Yicong and Sunkavalli, Kalyan and Shakhnarovich, Greg and Bi, Sai},
  journal={arXiv preprint arXiv:2311.06214},
  year={2023}
}

@article{animate124,
  title={Animate124: Animating one image to 4d dynamic scene},
  author={Zhao, Yuyang and Yan, Zhiwen and Xie, Enze and Hong, Lanqing and Li, Zhenguo and Lee, Gim Hee},
  journal={arXiv preprint arXiv:2311.14603},
  year={2023}
}

@article{4dfy,
  title={4d-fy: Text-to-4d generation using hybrid score distillation sampling},
  author={Bahmani, Sherwin and Skorokhodov, Ivan and Rong, Victor and Wetzstein, Gordon and Guibas, Leonidas and Wonka, Peter and Tulyakov, Sergey and Park, Jeong Joon and Tagliasacchi, Andrea and Lindell, David B},
  journal={arXiv preprint arXiv:2311.17984},
  year={2023}
}

@article{alignyg,
  title={Align your gaussians: Text-to-4d with dynamic 3d gaussians and composed diffusion models},
  author={Ling, Huan and Kim, Seung Wook and Torralba, Antonio and Fidler, Sanja and Kreis, Karsten},
  journal={arXiv preprint arXiv:2312.13763},
  year={2023}
}

@article{dreamgaussian4d,
  title={Dreamgaussian4d: Generative 4d gaussian splatting},
  author={Ren, Jiawei and Pan, Liang and Tang, Jiaxiang and Zhang, Chi and Cao, Ang and Zeng, Gang and Liu, Ziwei},
  journal={arXiv preprint arXiv:2312.17142},
  year={2023}
}

@inproceedings{sc4d,
  title={Sc4d: Sparse-controlled video-to-4d generation and motion transfer},
  author={Wu, Zijie and Yu, Chaohui and Jiang, Yanqin and Cao, Chenjie and Wang, Fan and Bai, Xiang},
  booktitle={European Conference on Computer Vision},
  pages={361--379},
  year={2024},
  organization={Springer}
}

@article{animate3d,
  title={Animate3d: Animating any 3d model with multi-view video diffusion},
  author={Jiang, Yanqin and Yu, Chaohui and Cao, Chenjie and Wang, Fan and Hu, Weiming and Gao, Jin},
  journal={arXiv preprint arXiv:2407.11398},
  year={2024}
}

@article{consistent4d,
  title={Consistent4d: Consistent 360 $\{$$\backslash$deg$\}$ dynamic object generation from monocular video},
  author={Jiang, Yanqin and Zhang, Li and Gao, Jin and Hu, Weimin and Yao, Yao},
  journal={arXiv preprint arXiv:2311.02848},
  year={2023}
}

@article{diffusion4d,
  title={Diffusion4d: Fast spatial-temporal consistent 4d generation via video diffusion models},
  author={Liang, Hanwen and Yin, Yuyang and Xu, Dejia and Liang, Hanxue and Wang, Zhangyang and Plataniotis, Konstantinos N and Zhao, Yao and Wei, Yunchao},
  journal={arXiv preprint arXiv:2405.16645},
  year={2024}
}

@article{l4gm,
  title={L4gm: Large 4d gaussian reconstruction model},
  author={Ren, Jiawei and Xie, Cheng and Mirzaei, Ashkan and Kreis, Karsten and Liu, Ziwei and Torralba, Antonio and Fidler, Sanja and Kim, Seung Wook and Ling, Huan and others},
  journal={Advances in Neural Information Processing Systems},
  volume={37},
  pages={56828--56858},
  year={2025}
}

@inproceedings{dream3d,
  title={Dream3d: Zero-shot text-to-3d synthesis using 3d shape prior and text-to-image diffusion models},
  author={Xu, Jiale and Wang, Xintao and Cheng, Weihao and Cao, Yan-Pei and Shan, Ying and Qie, Xiaohu and Gao, Shenghua},
  booktitle={Proceedings of the IEEE/CVF Conference on Computer Vision and Pattern Recognition},
  pages={20908--20918},
  year={2023}
}

@article{zero123++,
  title={Zero123++: a single image to consistent multi-view diffusion base model},
  author={Shi, Ruoxi and Chen, Hansheng and Zhang, Zhuoyang and Liu, Minghua and Xu, Chao and Wei, Xinyue and Chen, Linghao and Zeng, Chong and Su, Hao},
  journal={arXiv preprint arXiv:2310.15110},
  year={2023}
}

@inproceedings{richdreamer,
  title={Richdreamer: A generalizable normal-depth diffusion model for detail richness in text-to-3d},
  author={Qiu, Lingteng and Chen, Guanying and Gu, Xiaodong and Zuo, Qi and Xu, Mutian and Wu, Yushuang and Yuan, Weihao and Dong, Zilong and Bo, Liefeng and Han, Xiaoguang},
  booktitle={Proceedings of the IEEE/CVF conference on computer vision and pattern recognition},
  pages={9914--9925},
  year={2024}
}

@inproceedings{lgm,
  title={Lgm: Large multi-view gaussian model for high-resolution 3d content creation},
  author={Tang, Jiaxiang and Chen, Zhaoxi and Chen, Xiaokang and Wang, Tengfei and Zeng, Gang and Liu, Ziwei},
  booktitle={European Conference on Computer Vision},
  pages={1--18},
  year={2024},
  organization={Springer}
}

@article{lrm,
  title={Lrm: Large reconstruction model for single image to 3d},
  author={Hong, Yicong and Zhang, Kai and Gu, Jiuxiang and Bi, Sai and Zhou, Yang and Liu, Difan and Liu, Feng and Sunkavalli, Kalyan and Bui, Trung and Tan, Hao},
  journal={arXiv preprint arXiv:2311.04400},
  year={2023}
}

@article{meshlrm,
  title={MeshLRM: Large Reconstruction Model for High-Quality Meshes},
  author={Wei, Xinyue and Zhang, Kai and Bi, Sai and Tan, Hao and Luan, Fujun and Deschaintre, Valentin and Sunkavalli, Kalyan and Su, Hao and Xu, Zexiang},
  journal={arXiv preprint arXiv:2404.12385},
  year={2024}
}

@article{12345,
  title={One-2-3-45: Any single image to 3d mesh in 45 seconds without per-shape optimization},
  author={Liu, Minghua and Xu, Chao and Jin, Haian and Chen, Linghao and Varma T, Mukund and Xu, Zexiang and Su, Hao},
  journal={Advances in Neural Information Processing Systems},
  volume={36},
  pages={22226--22246},
  year={2023}
}

@inproceedings{12345++,
  title={One-2-3-45++: Fast single image to 3d objects with consistent multi-view generation and 3d diffusion},
  author={Liu, Minghua and Shi, Ruoxi and Chen, Linghao and Zhang, Zhuoyang and Xu, Chao and Wei, Xinyue and Chen, Hansheng and Zeng, Chong and Gu, Jiayuan and Su, Hao},
  booktitle={Proceedings of the IEEE/CVF conference on computer vision and pattern recognition},
  pages={10072--10083},
  year={2024}
}

@article{instantmesh,
  title={Instantmesh: Efficient 3d mesh generation from a single image with sparse-view large reconstruction models},
  author={Xu, Jiale and Cheng, Weihao and Gao, Yiming and Wang, Xintao and Gao, Shenghua and Shan, Ying},
  journal={arXiv preprint arXiv:2404.07191},
  year={2024}
}

@inproceedings{grm,
  title={Grm: Large gaussian reconstruction model for efficient 3d reconstruction and generation},
  author={Xu, Yinghao and Shi, Zifan and Yifan, Wang and Chen, Hansheng and Yang, Ceyuan and Peng, Sida and Shen, Yujun and Wetzstein, Gordon},
  booktitle={European Conference on Computer Vision},
  pages={1--20},
  year={2024},
  organization={Springer}
}

@inproceedings{gs-lrm,
  title={Gs-lrm: Large reconstruction model for 3d gaussian splatting},
  author={Zhang, Kai and Bi, Sai and Tan, Hao and Xiangli, Yuanbo and Zhao, Nanxuan and Sunkavalli, Kalyan and Xu, Zexiang},
  booktitle={European Conference on Computer Vision},
  pages={1--19},
  year={2024},
  organization={Springer}
}

@article{3dgs,
  title={3d gaussian splatting for real-time radiance field rendering.},
  author={Kerbl, Bernhard and Kopanas, Georgios and Leimk{\"u}hler, Thomas and Drettakis, George},
  journal={ACM Trans. Graph.},
  volume={42},
  number={4},
  pages={139--1},
  year={2023}
}

@article{nerf,
  title={Nerf: Representing scenes as neural radiance fields for view synthesis},
  author={Mildenhall, Ben and Srinivasan, Pratul P and Tancik, Matthew and Barron, Jonathan T and Ramamoorthi, Ravi and Ng, Ren},
  journal={Communications of the ACM},
  volume={65},
  number={1},
  pages={99--106},
  year={2021},
  publisher={ACM New York, NY, USA}
}

@article{svd,
  title={Stable video diffusion: Scaling latent video diffusion models to large datasets},
  author={Blattmann, Andreas and Dockhorn, Tim and Kulal, Sumith and Mendelevitch, Daniel and Kilian, Maciej and Lorenz, Dominik and Levi, Yam and English, Zion and Voleti, Vikram and Letts, Adam and others},
  journal={arXiv preprint arXiv:2311.15127},
  year={2023}
}

@article{animatediff,
  title={Animatediff: Animate your personalized text-to-image diffusion models without specific tuning},
  author={Guo, Yuwei and Yang, Ceyuan and Rao, Anyi and Liang, Zhengyang and Wang, Yaohui and Qiao, Yu and Agrawala, Maneesh and Lin, Dahua and Dai, Bo},
  journal={arXiv preprint arXiv:2307.04725},
  year={2023}
}

@inproceedings{stag4d,
  title={Stag4d: Spatial-temporal anchored generative 4d gaussians},
  author={Zeng, Yifei and Jiang, Yanqin and Zhu, Siyu and Lu, Yuanxun and Lin, Youtian and Zhu, Hao and Hu, Weiming and Cao, Xun and Yao, Yao},
  booktitle={European Conference on Computer Vision},
  pages={163--179},
  year={2024},
  organization={Springer}
}

@article{dreamgaussian,
  title={Dreamgaussian: Generative gaussian splatting for efficient 3d content creation},
  author={Tang, Jiaxiang and Ren, Jiawei and Zhou, Hang and Liu, Ziwei and Zeng, Gang},
  journal={arXiv preprint arXiv:2309.16653},
  year={2023}
}

@article{gaussiandreamer,
  title={Gaussiandreamer: Fast generation from text to 3d gaussian splatting with point cloud priors},
  author={Yi, Taoran and Fang, Jiemin and Wu, Guanjun and Xie, Lingxi and Zhang, Xiaopeng and Liu, Wenyu and Tian, Qi and Wang, Xinggang},
  journal={arXiv preprint arXiv:2310.08529},
  year={2023}
}

@article{gsgen,
  title={Text-to-3d using gaussian splatting},
  author={Chen, Zilong and Wang, Feng and Liu, Huaping},
  journal={arXiv preprint arXiv:2309.16585},
  year={2023}
}

@article{4dgen,
  title={4dgen: Grounded 4d content generation with spatial-temporal consistency},
  author={Yin, Yuyang and Xu, Dejia and Wang, Zhangyang and Zhao, Yao and Wei, Yunchao},
  journal={arXiv preprint arXiv:2312.17225},
  year={2023}
}

@article{efficient4d,
  title={Fast Dynamic 3D Object Generation from a Single-view Video},
  author={Pan, Zijie and Yang, Zeyu and Zhu, Xiatian and Zhang, Li},
  journal={arXiv preprint arXiv:2401.08742},
  year={2024}
}

@article{4diffusion,
  title={4diffusion: Multi-view video diffusion model for 4d generation},
  author={Zhang, Haiyu and Chen, Xinyuan and Wang, Yaohui and Liu, Xihui and Wang, Yunhong and Qiao, Yu},
  journal={Advances in Neural Information Processing Systems},
  volume={37},
  pages={15272--15295},
  year={2025}
}

@article{vividzoo,
  title={Vivid-zoo: Multi-view video generation with diffusion model},
  author={Li, Bing and Zheng, Cheng and Zhu, Wenxuan and Mai, Jinjie and Zhang, Biao and Wonka, Peter and Ghanem, Bernard},
  journal={Advances in Neural Information Processing Systems},
  volume={37},
  pages={62189--62222},
  year={2025}
}

@misc{zeroscope,
    title = {Zeroscope text-to-video model},
    author = {Cerspense},
    year = {2023},
    howpublished = {\url{https://huggingface.co/cerspense/zeroscope_v2_576w}},
    note = {Accessed: 2023-10-31}
}

@misc{sketchfab,
    title = {Sketchfab},
    author = {Sketchfab},
    year = {2024},
    howpublished = {\url{https://sketchfab.com/}},
    note = {Accessed: 2024-05-21}
}

@article{modelscope,
  title={Modelscope text-to-video technical report},
  author={Wang, Jiuniu and Yuan, Hangjie and Chen, Dayou and Zhang, Yingya and Wang, Xiang and Zhang, Shiwei},
  journal={arXiv preprint arXiv:2308.06571},
  year={2023}
}

@article{smpl,
  title={SMPL: A skinned multi-person linear model},
  author={Loper, Matthew and Mahmood, Naureen and Romero, Javier and Pons-Moll, Gerard and Black, Michael J},
  journal={Seminal Graphics Papers: Pushing the Boundaries, Volume 2},
  pages={851--866},
  year={2023}
}

@article{3dshape2vecset,
  title={3dshape2vecset: A 3d shape representation for neural fields and generative diffusion models},
  author={Zhang, Biao and Tang, Jiapeng and Niessner, Matthias and Wonka, Peter},
  journal={ACM Transactions On Graphics (TOG)},
  volume={42},
  number={4},
  pages={1--16},
  year={2023},
  publisher={ACM New York, NY, USA}
}

@inproceedings{motion2vecsets,
  title={Motion2vecsets: 4d latent vector set diffusion for non-rigid shape reconstruction and tracking},
  author={Cao, Wei and Luo, Chang and Zhang, Biao and Nie{\ss}ner, Matthias and Tang, Jiapeng},
  booktitle={Proceedings of the IEEE/CVF conference on computer vision and pattern recognition},
  pages={20496--20506},
  year={2024}
}

@conference{AMASS,
  title           = {{AMASS}: Archive of Motion Capture as Surface Shapes},
  author          = {Mahmood, Naureen and Ghorbani, Nima and Troje, Nikolaus F. and Pons-Moll, Gerard and Black, Michael J.},
  booktitle       = {International Conference on Computer Vision},
  pages           = {5442--5451},
  month           = oct,
  year            = {2019},
  month_numeric   = {10}
}

@inproceedings{dt4d,
  title={4dcomplete: Non-rigid motion estimation beyond the observable surface},
  author={Li, Yang and Takehara, Hikari and Taketomi, Takafumi and Zheng, Bo and Nie{\ss}ner, Matthias},
  booktitle={Proceedings of the IEEE/CVF International Conference on Computer Vision},
  pages={12706--12716},
  year={2021}
}

@article{motiondiffuse,
  title={Motiondiffuse: Text-driven human motion generation with diffusion model},
  author={Zhang, Mingyuan and Cai, Zhongang and Pan, Liang and Hong, Fangzhou and Guo, Xinying and Yang, Lei and Liu, Ziwei},
  journal={IEEE transactions on pattern analysis and machine intelligence},
  volume={46},
  number={6},
  pages={4115--4128},
  year={2024},
  publisher={IEEE}
}

@inproceedings{realistichumanmt,
  title={Realistic human motion generation with cross-diffusion models},
  author={Ren, Zeping and Huang, Shaoli and Li, Xiu},
  booktitle={European Conference on Computer Vision},
  pages={345--362},
  year={2024},
  organization={Springer}
}

@article{humanmdm,
  title={Human motion diffusion model},
  author={Tevet, Guy and Raab, Sigal and Gordon, Brian and Shafir, Yonatan and Cohen-Or, Daniel and Bermano, Amit H},
  journal={arXiv preprint arXiv:2209.14916},
  year={2022}
}

@inproceedings{physdiff,
  title={Physdiff: Physics-guided human motion diffusion model},
  author={Yuan, Ye and Song, Jiaming and Iqbal, Umar and Vahdat, Arash and Kautz, Jan},
  booktitle={Proceedings of the IEEE/CVF international conference on computer vision},
  pages={16010--16021},
  year={2023}
}

@article{rf,
  title={Flow straight and fast: Learning to generate and transfer data with rectified flow},
  author={Liu, Xingchao and Gong, Chengyue and Liu, Qiang},
  journal={arXiv preprint arXiv:2209.03003},
  year={2022}
}

@article{zhang2025lpm,
  title={LPM: Efficient 3D Content Creation from Single Image by Large-Scale Partial 3D Modeling},
  author={Zhang, Yisu and Yu, Chaohui and Wang, Fan and Zhu, Jianke},
  journal={IEEE Transactions on Circuits and Systems for Video Technology},
  year={2025},
  publisher={IEEE}
}

@inproceedings{dit,
  title={Scalable diffusion models with transformers},
  author={Peebles, William and Xie, Saining},
  booktitle={Proceedings of the IEEE/CVF international conference on computer vision},
  pages={4195--4205},
  year={2023}
}

@article{qwen25,
  title={Qwen2. 5-VL Technical Report},
  author={Bai, Shuai and Chen, Keqin and Liu, Xuejing and Wang, Jialin and Ge, Wenbin and Song, Sibo and Dang, Kai and Wang, Peng and Wang, Shijie and Tang, Jun and others},
  journal={arXiv preprint arXiv:2502.13923},
  year={2025}
}

@inproceedings{vbench,
  title={Vbench: Comprehensive benchmark suite for video generative models},
  author={Huang, Ziqi and He, Yinan and Yu, Jiashuo and Zhang, Fan and Si, Chenyang and Jiang, Yuming and Zhang, Yuanhan and Wu, Tianxing and Jin, Qingyang and Chanpaisit, Nattapol and others},
  booktitle={Proceedings of the IEEE/CVF Conference on Computer Vision and Pattern Recognition},
  pages={21807--21818},
  year={2024}
}

@inproceedings{arap,
  title={As-rigid-as-possible surface modeling},
  author={Sorkine, Olga and Alexa, Marc},
  booktitle={Symposium on Geometry processing},
  volume={4},
  pages={109--116},
  year={2007},
  organization={Citeseer}
}

@article{hunyuan1.0,
  title={Hunyuan3d 1.0: A unified framework for text-to-3d and image-to-3d generation},
  author={Yang, Xianghui and Shi, Huiwen and Zhang, Bowen and Yang, Fan and Wang, Jiacheng and Zhao, Hongxu and Liu, Xinhai and Wang, Xinzhou and Lin, Qingxiang and Yu, Jiaao and others},
  journal={arXiv preprint arXiv:2411.02293},
  year={2024}
}

@article{hunyuan2.0,
  title={Hunyuan3d 2.0: Scaling diffusion models for high resolution textured 3d assets generation},
  author={Zhao, Zibo and Lai, Zeqiang and Lin, Qingxiang and Zhao, Yunfei and Liu, Haolin and Yang, Shuhui and Feng, Yifei and Yang, Mingxin and Zhang, Sheng and Yang, Xianghui and others},
  journal={arXiv preprint arXiv:2501.12202},
  year={2025}
}

@article{animateanymesh,
  title={AnimateAnyMesh: A Feed-Forward 4D Foundation Model for Text-Driven Universal Mesh Animation},
  author={Wu, Zijie and Yu, Chaohui and Wang, Fan and Bai, Xiang},
  journal={arXiv preprint arXiv:2506.09982},
  year={2025}
}

@inproceedings{trellis,
  title={Structured 3d latents for scalable and versatile 3d generation},
  author={Xiang, Jianfeng and Lv, Zelong and Xu, Sicheng and Deng, Yu and Wang, Ruicheng and Zhang, Bowen and Chen, Dong and Tong, Xin and Yang, Jiaolong},
  booktitle={Proceedings of the Computer Vision and Pattern Recognition Conference},
  pages={21469--21480},
  year={2025}
}

@article{triposg,
  title={Triposg: High-fidelity 3d shape synthesis using large-scale rectified flow models},
  author={Li, Yangguang and Zou, Zi-Xin and Liu, Zexiang and Wang, Dehu and Liang, Yuan and Yu, Zhipeng and Liu, Xingchao and Guo, Yuan-Chen and Liang, Ding and Ouyang, Wanli and others},
  journal={arXiv preprint arXiv:2502.06608},
  year={2025}
}

@article{sparseflex,
  title={Sparseflex: High-resolution and arbitrary-topology 3d shape modeling},
  author={He, Xianglong and Zou, Zi-Xin and Chen, Chia-Hao and Guo, Yuan-Chen and Liang, Ding and Yuan, Chun and Ouyang, Wanli and Cao, Yan-Pei and Li, Yangguang},
  journal={arXiv preprint arXiv:2503.21732},
  year={2025}
}

@article{hi3dgen,
  title={Hi3dgen: High-fidelity 3d geometry generation from images via normal bridging},
  author={Ye, Chongjie and Wu, Yushuang and Lu, Ziteng and Chang, Jiahao and Guo, Xiaoyang and Zhou, Jiaqing and Zhao, Hao and Han, Xiaoguang},
  journal={arXiv preprint arXiv:2503.22236},
  volume={3},
  pages={2},
  year={2025}
}

@article{direct3d,
  title={Direct3d-s2: Gigascale 3d generation made easy with spatial sparse attention},
  author={Wu, Shuang and Lin, Youtian and Zhang, Feihu and Zeng, Yifei and Yang, Yikang and Bao, Yajie and Qian, Jiachen and Zhu, Siyu and Cao, Xun and Torr, Philip and others},
  journal={arXiv preprint arXiv:2505.17412},
  year={2025}
}

@article{wan,
  title={Wan: Open and advanced large-scale video generative models},
  author={Wan, Team and Wang, Ang and Ai, Baole and Wen, Bin and Mao, Chaojie and Xie, Chen-Wei and Chen, Di and Yu, Feiwu and Zhao, Haiming and Yang, Jianxiao and others},
  journal={arXiv preprint arXiv:2503.20314},
  year={2025}
}

@article{rope,
  title={Roformer: Enhanced transformer with rotary position embedding},
  author={Su, Jianlin and Ahmed, Murtadha and Lu, Yu and Pan, Shengfeng and Bo, Wen and Liu, Yunfeng},
  journal={Neurocomputing},
  volume={568},
  pages={127063},
  year={2024},
  publisher={Elsevier}
}

@article{umap,
  title={Umap: Uniform manifold approximation and projection for dimension reduction},
  author={McInnes, Leland and Healy, John and Melville, James},
  journal={arXiv preprint arXiv:1802.03426},
  year={2018}
}

@inproceedings{motioneditor,
  title={MotionEditor: Editing Video Motion via Content-Aware Diffusion},
  author={Tu, Shuyuan and Dai, Qi and Cheng, Zhi-Qi and Hu, Han and Han, Xintong and Wu, Zuxuan and Jiang, Yu-Gang},
  booktitle={Proceedings of the IEEE/CVF Conference on Computer Vision and Pattern Recognition (CVPR)},
  pages={7882--7891},
  year={2024}
}

@inproceedings{motionfollower,
  title={MotionFollower: Editing Video Motion via Lightweight Score-Guided Diffusion},
  author={Tu, Shuyuan and Dai, Qi and Cheng, Zhi-Qi and Han, Xintong and Wu, Zuxuan and Jiang, Yu-Gang},
  booktitle={Proceedings of the IEEE/CVF International Conference on Computer Vision (ICCV)},
  year={2025}
}

@inproceedings{stableanimator,
  title={StableAnimator: High-Quality Identity-Preserving Human Image Animation},
  author={Tu, Shuyuan and Xing, Zhen and Han, Xintong and Cheng, Zhi-Qi and Dai, Qi and Luo, Chong and Wu, Zuxuan},
  booktitle={Proceedings of the IEEE/CVF Conference on Computer Vision and Pattern Recognition (CVPR)},
  pages={21096--21106},
  year={2025}
}

@article{stableanimatorpp,
  title={StableAnimator++: Overcoming Pose Misalignment and Face Distortion for Human Image Animation},
  author={Tu, Shuyuan and Xing, Zhen and Han, Xintong and Cheng, Zhi-Qi and Dai, Qi and Luo, Chong and Wu, Zuxuan},
  journal={arXiv preprint arXiv:2507.15064},
  year={2025}
}

\begin{IEEEbiography}[{\includegraphics[width=1in,height=1.25in,clip,keepaspectratio]{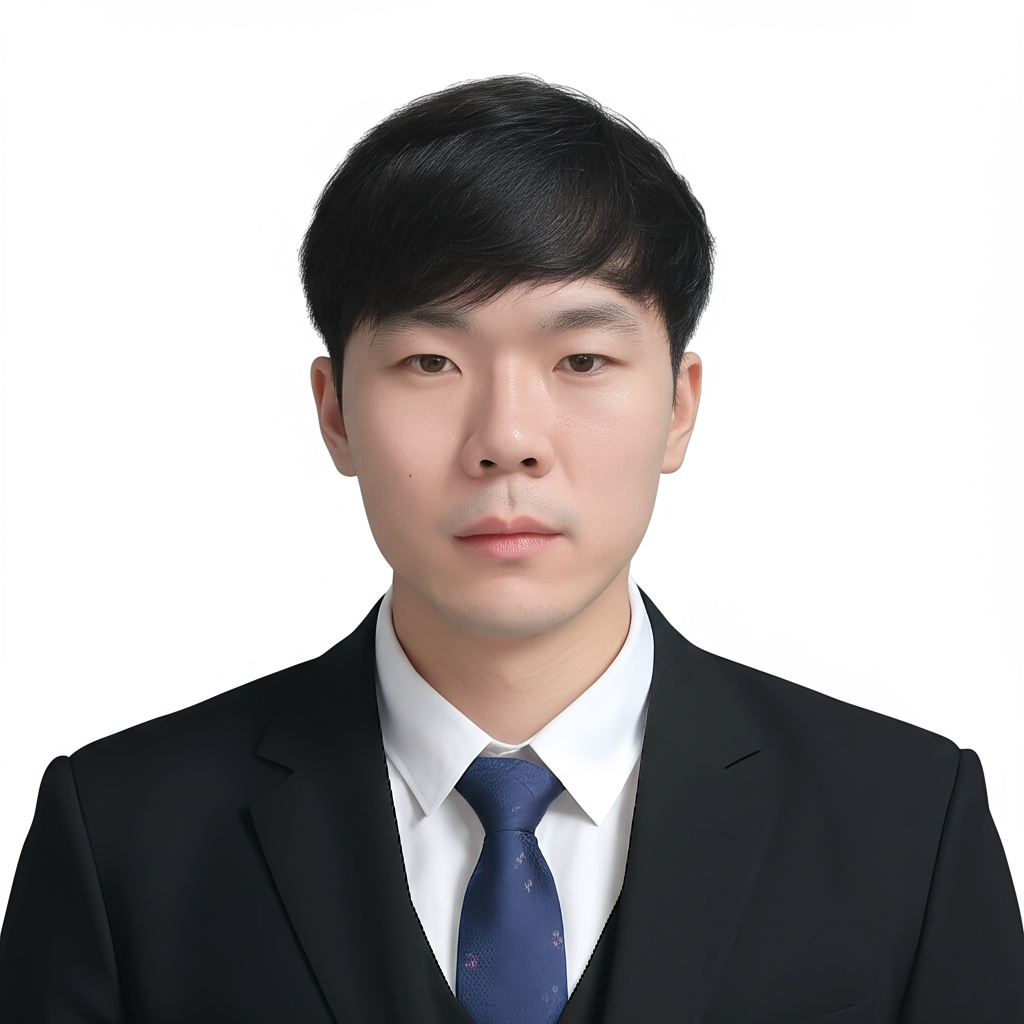}}]{Zijie Wu}
received his B.S and M.S. degrees from the School of Electronic Information and Communications, Huazhong University of Science and Technology~(HUST), Wuhan, China, in 2023. He is currently a Ph.D. candidate with the School of Artificial Intelligence and Automation, HUST. His research interests focus on 2D/3D/4D AIGC, especially 4D generation and motion synthesis.
\end{IEEEbiography}

\begin{IEEEbiography}[{\includegraphics[width=1in,height=1.25in,clip,keepaspectratio]{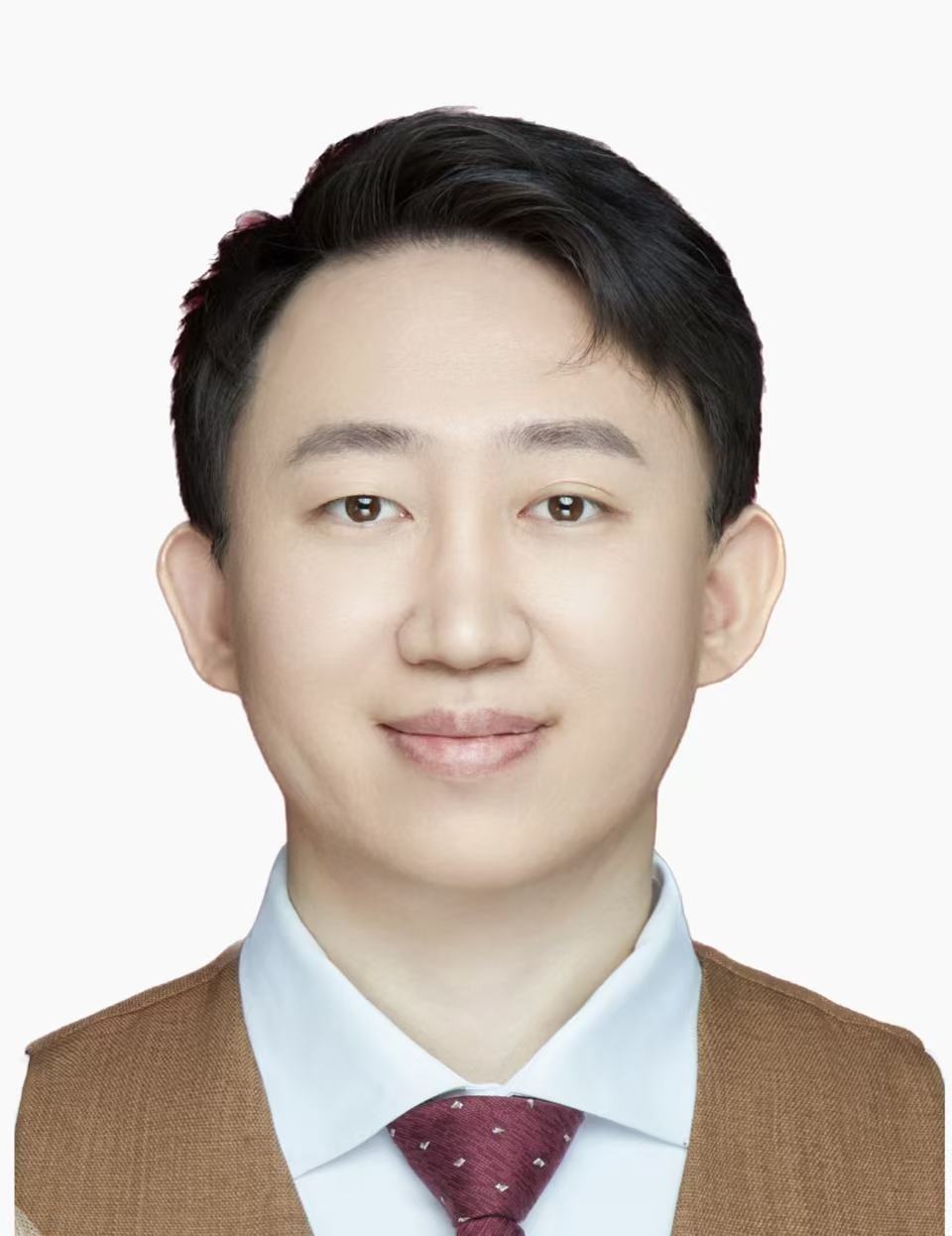}}]{Chaohui Yu}
received the bachelor’s degree from Shandong University in 2017, and the master’s degree from the Institute of Computing Technology (ICT), Chinese Academy of Sciences, Beijing in 2020. He is currently a researcher of DAMO Academy, Alibaba Group, Beijing, China. His research interests include computer vision and related applications.
\end{IEEEbiography}

\begin{IEEEbiography}
[{\includegraphics[width=1in,height=1.25in,clip,keepaspectratio]{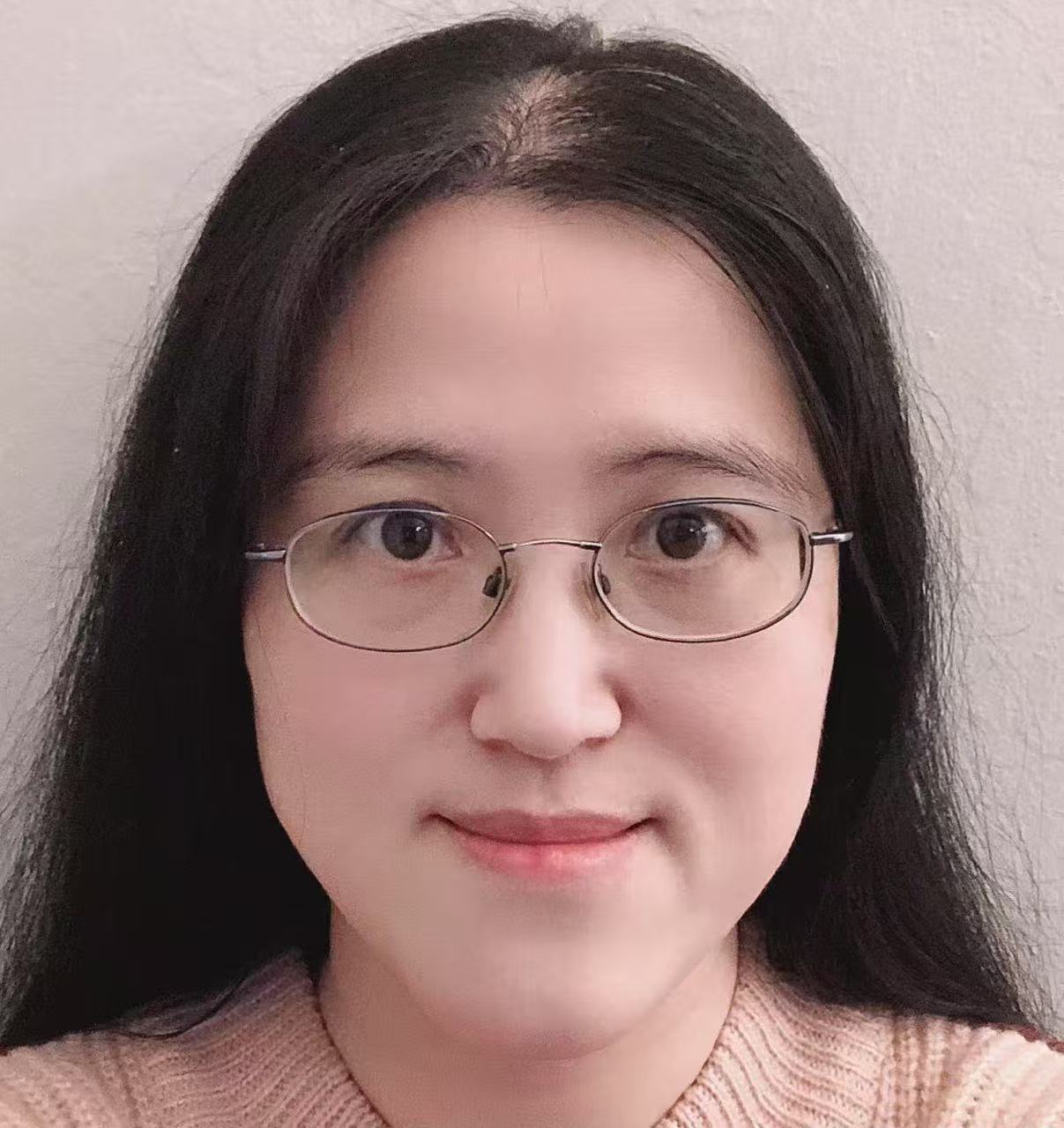}}]{Fan Wang}
received the B.S. and M.S. degrees from the Department of Automation, Tsinghua University, Beijing, China, and the Ph.D. degree from the Department of Electrical Engineering, Stanford University, Stanford, CA, USA. She is currently working as a Distinguished Scientist with Damo Academy, Alibaba Group. Her research interests include generative AI, 3D vision, and multimodal learning.
\end{IEEEbiography}

\begin{IEEEbiography}[{\includegraphics[width=1in,height=1.25in,clip,keepaspectratio]{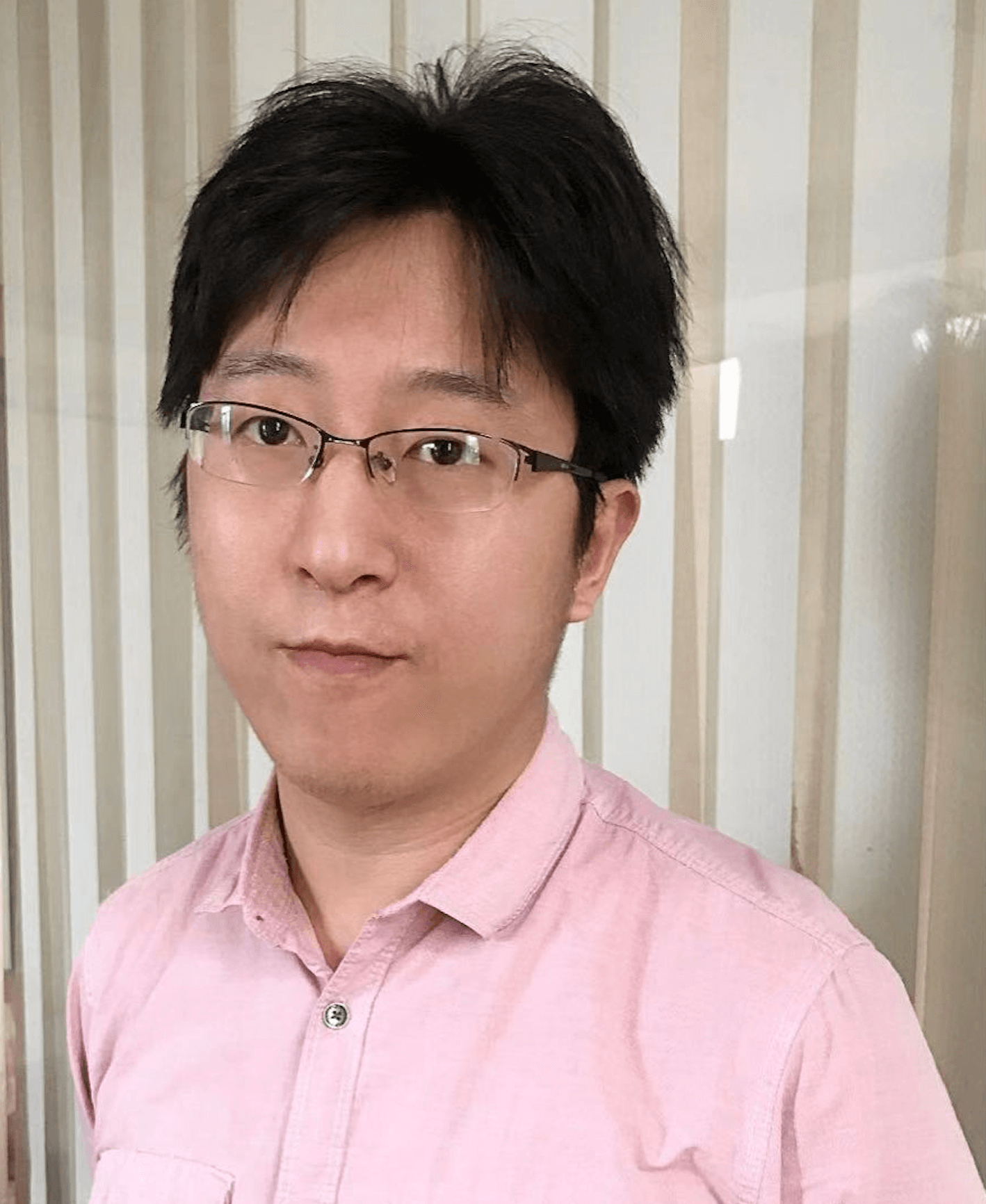}}]{Xiang Bai}
received his B.S., M.S., and Ph.D. degrees from the Huazhong University of Science and Technology~(HUST), Wuhan, China, in 2003, 2005, and 2009, respectively, all in electronics and information engineering. He is currently a Professor and the Dean of the School of Software Engineering, HUST. He is also the Vice-director of the National Center of Anti-Counterfeiting Technology, HUST. His research interests include object recognition, shape analysis, scene text recognition, and intelligent systems. He serves as an Associate Editor for IEEE Transactions on Pattern Analysis and Machine Intelligence, Pattern Recognition, Pattern Recognition Letters, Neurocomputing, and Frontiers of Computer Science. He is a Fellow of the IEEE.
\end{IEEEbiography}

\end{document}